\documentclass{article}

\usepackage{microtype}
\usepackage{graphicx}
\usepackage{subfigure}
\usepackage{booktabs} % for professional tables
\usepackage{dcolumn}
\newcolumntype{d}[1]{D{.}{.}{#1}}

\usepackage{xurl}
\usepackage{hyperref}
\usepackage{float}
\usepackage{seqsplit}
\usepackage{siunitx}

\usepackage[accepted]{icml2026} 

\usepackage{amsmath}
\usepackage{amssymb}
\usepackage{mathtools}
\usepackage{amsthm}
\usepackage{enumitem}      % For itemize with [left=1pt] option

\usepackage[capitalize,noabbrev]{cleveref}

\usepackage{ragged2e}

\theoremstyle{plain}

\theoremstyle{definition}

\theoremstyle{remark}

\icmltitlerunning{On the Limits of LLM Adaptability: Impact of Model-Internalized Priors on Annotation Task Performance}

\begin{document}

\twocolumn[
\icmltitle{On the Limits of LLM Adaptability: Impact of Model-Internalized Priors on Annotation Task Performance    }

\icmlsetsymbol{equal}{*}

\begin{icmlauthorlist}
\icmlauthor{Etienne Casanova}{equal,Caltech}
\icmlauthor{Rafal Kocielnik}{equal,Caltech}
\icmlauthor{R. Michael Alvarez}{Caltech}
\end{icmlauthorlist}

\icmlaffiliation{Caltech}{California Institute of Technology, Pasadena, CA, USA}

\icmlcorrespondingauthor{Etienne Casanova}{ecasanov@caltech.edu}
\icmlcorrespondingauthor{Rafal Kocielnik}{rafalko@caltech.edu}

\icmlkeywords{Large Language Models, Model-Internalized Priors, Zero-Shot Learning, Prompt Steerability, Annotation Reliability, Model Controllability}

\vskip 0.3in
]

\printAffiliationsAndNotice{\icmlEqualContribution}

\begin{abstract}
Large Language Models (LLMs) are increasingly used for zero-shot annotation and LLM-as-a-judge tasks, yet their reliability hinges on how model-internalized priors interact with user-provided instructions. 
We investigate three dimensions of this interaction: (1) how an LLM's familiarity with data and task definitions relates to performance, (2) whether additional information in prompts can correct zero-shot errors (``decision stickiness''), and (3) model susceptibility to misaligned task definitions. 
We introduce \textbf{\emph{Definition-Specific Familiarity (DSF)}}, which measures alignment between a model's elicited concept and the target definition. 
Across nine LLMs and six toxicity datasets (five primary datasets plus an additional robustness dataset), DSF predicts annotation performance after controlling for dataset identity (partial $r=+0.41$). 
This association remains positive across all prompting conditions tested. 
In contrast, three common text-memorization metrics show no positive association. 
We show that prompting has limited corrective power: only 34.8\% of zero-shot errors are corrected by additional instructions or examples, with high-confidence errors especially persistent. 
Misaligned definitions systematically shift predictions without reducing reported confidence, making confidence unreliable for detecting definition--policy mismatch. 
Together, these findings establish definition alignment as a practical model-selection criterion and show that better prompting alone cannot substitute for validating model--policy fit.
\end{abstract}

\begin{figure*}[t]
\centering
    \includegraphics[width=\textwidth]{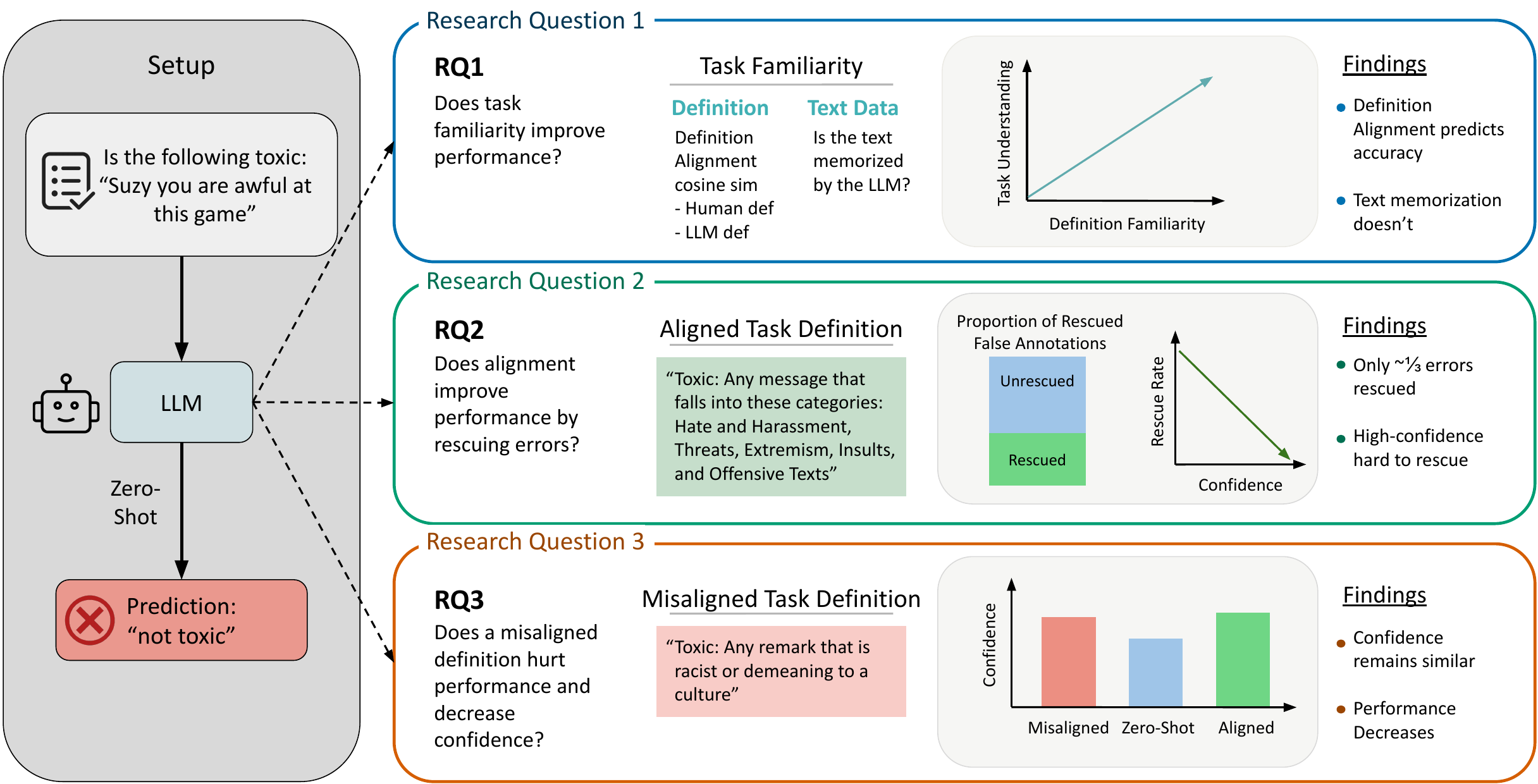}
    \vspace{-1.5em}
    \caption{\textbf{Study overview and research questions.} \emph{Left:} zero-shot annotation setup: given an input text and a user-provided task definition, the LLM produces a label prediction. \emph{Right:} we study three facets of the interaction between model-internalized task concepts and user instructions. \textbf{RQ1} tests whether task familiarity correlates with performance, contrasting \emph{definition alignment} (Definition-Specific Familiarity; DSF) with \emph{text memorization} metrics (ROUGE-L, BERTScore, embedding similarity; Table~\ref{tab:familiarity_correlations}). \textbf{RQ2} measures steerability under an aligned definition by asking whether zero-shot errors can be \emph{rescued} (rescue rate), and how this depends on confidence. \textbf{RQ3} probes susceptibility to \emph{misaligned} definitions, testing how performance and confidence change when the provided definition is incorrect. Insets summarize the main empirical findings for each question.}
    \vspace{-8pt}
\label{fig:overview}
\end{figure*}

\section{Introduction}
Large Language Models (LLMs) are increasingly deployed for zero-shot annotation \cite{gilardi2023chatgpt} and rubric-based evaluation (LLM-as-a-judge) tasks \cite{zheng2023judging,liu2023geval}. These use cases are attractive: users provide a task definition via prompt, and the model annotates at scale without labeled training data \cite{gilardi2023chatgpt,brown2020language}. The implicit assumption is that the user's definition governs the model's behavior.

But LLMs are not blank slates. Through pre-training on web-scale corpora \cite{brown2020language}, followed by instruction tuning and reinforcement learning from human feedback (RLHF), models develop implicit understandings of common concepts and tasks. When a user asks an LLM to detect ``toxic'' content, the model has already been shaped by millions of documents and human-rated examples that discuss, define, and exemplify toxicity, and this exposure shapes an internal concept of the task. Critically, this internalized concept may not align with the user's intended definition: toxicity is operationalized differently across applications, as identity attacks \citep{borkan2019nuanced}, disruptive behavior in online gaming \citep{kocielnik2024challenges, kocielnik2025online}, or as speech targeting protected groups \citep{elsherief2018hatelingo}. An LLM anchored to a particular understanding may not simultaneously match all of these definitions, and its internalized priors may constrain how much user instructions can steer its behavior and alignment with the user's intended interpretation \cite{min2022rethinking}. This disconnect between a model's internalized concepts and user-specified definitions represents a fundamental alignment challenge.

When the user's task definition conflicts with the model's internalized concept, which one governs behavior? Prior work has shown that LLMs exhibit biases reflecting the demographics and viewpoints in their training data --- both at the pre-training stage and after instruction tuning or RLHF \cite{nadeem2021stereoset,nangia2020crows,parrish2022bbq,santurkar2023whose}. Prompting and reasoning techniques can inadvertently amplify such latent biases \cite{shaikh2023second}. Less is known about the degree to which users can override a model's internalized annotation task understanding through prompting, or how to measure the alignment between a model's internal concept and a task definition.

We investigate this tension through three research questions (Figure~\ref{fig:overview}):

\vspace{-0.3em}
\begin{itemize}[leftmargin=1.5em, labelsep=0.3em, topsep=0pt, itemsep=2pt, parsep=0pt]
\vspace{-0.2em}

\item\textbf{RQ1 (Definition Fit and Model Selection):} Does alignment between a model's internalized concept and a task definition predict which model performs better on that task, beyond text memorization and dataset differences?
\item \textbf{RQ2 (Decision Stickiness):} To what extent can external knowledge or specialized prompting correct initial zero-shot errors, and how does model confidence relate to correctability? 
\item \textbf{RQ3 (Misalignment Susceptibility):} How do LLMs behave when given misaligned or incorrect task definitions, and can confidence scores detect such misalignment?

\end{itemize}
\vspace{-0.6em}

We evaluate nine LLMs on five primary toxicity datasets spanning social media, gaming, news, and forums, with three supplementary datasets providing targeted robustness and non-toxicity generalization checks. We make three contributions. First, we introduce \emph{Definition-Specific Familiarity (DSF)}, which compares a model's elicited concept with a dataset's definition. Across 54 model--dataset pairs, DSF predicts which models perform better within the same task after controlling for dataset identity (partial $r=+0.41$); three text-memorization measures provide no predictive positive signal. Second, we show that prompting is a limited correction mechanism: only 34.8\% of zero-shot errors are rescued by definitions, examples, or automated prompt refinement, and high-confidence errors are especially sticky. Third, misaligned definitions systematically redirect predictions without a corresponding confidence decrease, making confidence unreliable for detecting definition--policy mismatch. Together, these results show that model selection and policy validation cannot be replaced by better prompting or confidence thresholds alone. They extend work on steerability in generation \cite{chang2025steerability} by characterizing the limits of prompting as a control channel for discrete annotation decisions. Code and analysis pipelines are available at \url{https://github.com/etmaca5/llm-internalized-priors-for-annotation}.

%%%%%%%%% BACKgROUND  %%%%

\section{Background and Related Work}

\paragraph{LLM Steerability and Controllability.}
Recent work has formalized steerability as the degree to which a model's behavior can be shifted from its baseline via interventions such as prompting \cite{miehling2025promptsteerability}. \citet{chang2025steerability} introduce a multi-dimensional goal-space framework for measuring steerability in text generation, identifying three key failure modes: poor coverage (rare goals underrepresented), miscalibration (overshooting requests), and side effects (unintended changes to unspecified dimensions). Their evaluation reveals that current LLMs struggle with reliable steerability even under various interventions including prompt engineering and reinforcement learning. Our work addresses a complementary question: in \emph{annotation} tasks where the output is a discrete label rather than generated text, what determines whether a model can be steered toward a specific task definition? We find analogous failure modes: limited reachability (stubborn errors), miscalibration (flat confidence under misalignment), and side effects (rescue-corruption tradeoffs).

\vspace{-0pt}
\paragraph{LLMs as Annotators and Judges.}
LLMs are widely used for text annotation, sometimes outperforming crowd-workers \citep{gilardi2023chatgpt}. Their use as evaluators spans judging open-ended responses \citep{zheng2023judging} and models specifically fine-tuned for detailed rubric-based assessment \citep{prometheus2024}. Reliable interpretation of evaluation criteria is especially important in high-stakes settings, such as surgical training, where LLM-derived quality scores have been assessed against observed trainee and trainer responses \citep{kocielnik2026surgical}. Large-scale analyses of social phenomena likewise depend on interpretive judgments: prior work emphasizes aligning model interpretations with human definitions \citep{kocielnik2025prosocial}, while documented false positives illustrate how broadly related content can be mistaken for evidence satisfying stricter labeling criteria \citep{casanova2026truth}. More generally, configuration choices can substantially alter research conclusions, a problem termed ``LLM hacking'' \citep{baumann2025llmhacking}, and transferring examples across labeling definitions can help or hurt performance \citep{kocielnik2023label}. These findings motivate our investigation of whether alignment between an elicited model concept and the target definition predicts annotation performance.

\vspace{-0pt}
\paragraph{Model-Internalized Priors and Annotation Reliability.}
Modern LLMs develop strong internal priors across the full training pipeline (pre-training, instruction tuning, and RLHF/DPO) that can conflict with user-provided definitions. \citet{carlini2021extracting} demonstrate that LLMs memorize specific training examples, suggesting models have ingrained conceptions of common concepts like toxicity. Post-training alignment can amplify these effects: \citet{zhang2025verbalized} identify \emph{typicality bias} where annotators systematically favor familiar text during RLHF, causing models to anchor to ``typical'' examples rather than following nuanced definitions. \citet{min2022rethinking} show that in-context learning is often driven by format and label space rather than input-label mappings, suggesting models may not fully override their priors even with explicit examples. Prior work on benchmark contamination has developed metrics like Min-K\% Prob \cite{shi2024detecting}, BERTScore \cite{zhang2020bertscore},  and continuation-based detection \cite{golchin2024time}, but less is known about how internalized \emph{conceptual} priors (vs. text memorization) constrain steerability.

\vspace{-0pt}
\paragraph{Confidence and Calibration.}
Model accuracy alone can mask calibration failures critical for annotation reliability. Surveys of LLM confidence estimation \cite{geng2024surveyconfidence} highlight challenges with overconfidence and unreliable uncertainty quantification. Studies of verbalized uncertainty report overconfidence and sensitivity to reconsideration prompts \cite{xiong2024uncertainty,pawitan2024confidence}. More broadly, dissociation between self-reported traits and observed behavior cautions against treating model self-descriptions as direct behavioral evidence \cite{han2025personality}. While verbalized confidence can be better calibrated than token probabilities under RLHF \cite{tian2023justask}, prompt design strongly shapes reliability \cite{yang2024verbalizedconfidence}. We adopt verbalized confidence to enable applicability to closed-source models that do not expose token-level probability scores.

\vspace{-0pt}
\section{Methods}
We propose a framework to evaluate the interplay between the priors an LLM has internalized through training and tuning (familiarity) and its steerability via prompting (alignment) in the context of annotation tasks.

\vspace{-0pt}
\subsection{Familiarity Metrics.}
We distinguish between two types of familiarity that could influence annotation performance: \emph{text familiarity} (whether the model has memorized specific texts from the dataset) and \emph{definition familiarity} (whether the model's internal concept aligns with the task definition). These represent different mechanisms: text familiarity suggests performance via generation of memorized content, while definition familiarity suggests performance via conceptual task alignment.

\vspace{-0pt}
\paragraph{Text Familiarity.} Measures memorization by prompting the model to generate a continuation given a prefix comprising 40\% of the text instance, then computing \textbf{ROUGE-L F1} (based on longest common subsequence) \cite{lin2004rouge} between the generated continuation and the remaining ground truth text. This adapts the continuation-based contamination detection approach of \citet{golchin2024time} for continuous familiarity measurement: rather than using dataset-specific prompts for binary contamination detection, we use generic continuation prompts (see Appendix~\ref{app:prompts}) at temperatures 0.0 and 0.7, taking the maximum score. To ensure findings are robust to the choice of memorization proxy, we additionally compute \textbf{BERTScore F1} \cite{zhang2020bertscore} using DeBERTa-XL contextual token embeddings, and \textbf{embedding cosine similarity} using the same sentence encoders as DSF---so any difference between metrics reflects signal type (lexical vs.\ contextual vs.\ semantic reproduction) rather than encoder choice. High scores indicate that the model can reproduce text that it likely encountered during training.

\vspace{-0pt}
\paragraph{Definition-Specific Familiarity (DSF).} Quantifies alignment between a model's internal understanding of the target phenomenon (e.g., what constitutes toxicity) and the dataset's operational definition of that phenomenon. The intuition is analogous to human annotators: two annotators given the same definition may interpret it differently based on their prior conceptions and identities \citep{sap2022annotators, plank2022problem}. An annotator whose mental model aligns with the definition will perform better than one with a mismatched conception, regardless of whether they have seen the specific texts before.

To compute DSF, we prompt the model to explain its understanding of the target concept (e.g., ``In your own words, what makes content toxic?'') and measure semantic similarity between this explanation and the dataset's full definition using sentence embeddings. To remove sensitivity to any single embedding model, we compute DSF as the unweighted mean (\emph{consensus DSF}) across six diverse sentence encoders: MiniLM, MPNet, BGE-large, E5-large, Instructor-large, and OpenAI's \texttt{text-embedding-3-small}. Appendix~\ref{app:dsf_embeddings} shows that all six encoders produce positive partial correlations with accuracy and that pairwise DSF values are highly concordant ($r \geq 0.88$). DSF--accuracy correlations are computed across all nine models and six datasets (adding the Jigsaw Unintended Bias dataset~\citep{borkan2019nuanced} to the five primary datasets), yielding $N=54$ model--dataset pairs. Crucially, DSF can be measured using only the task definition and a handful of prompts, without requiring labeled data or full annotation runs, making it a practical diagnostic for selecting model--task pairings. This quantification captures the entire concept boundary including both positive and negative criteria. We also tested alternative variants incorporating domain-specific prompts and separate positive/negative class definitions, but these showed weaker correlation with model performance on a given task (see Appendix~\ref{app:familiarity}).

\vspace{-0pt}
\subsection{Steerability Metrics}

To quantify an LLM's ability to correct its errors when provided with better instructions (steerability), we define the \textbf{Rescue Rate}:
\begin{equation}
    \text{Rescue Rate} = P(\text{Correct} \mid \text{Prompted}, \text{Zero-Shot Wrong})
\end{equation}
This metric captures the reachability of correct behaviors via prompting. A low rescue rate indicates that many behaviors are unreachable through prompt-based steering alone. We define \textbf{decision stickiness} as the tendency for high-confidence errors to resist correction.

To investigate how LLMs respond to incorrect instructions (RQ3), we employ \emph{concept substitution}: providing a definition from a different dataset that operationalizes a related but semantically distinct concept. For example, when annotating a general toxicity dataset, we substitute definitions of hate speech (which requires identity-based targeting) for gaming toxicity (which includes any disruptive behavior). We test 6 misaligned conditions varying in \emph{definition scope}: narrow definitions like hate speech require identity-based targeting, while broad definitions like gaming toxicity capture any disruptive behavior. Full definitions are in Appendix~\ref{app:definitions}.

We measure four complementary metrics that decompose how prompting changes behavior relative to zero-shot: \textbf{Change Rate} (fraction of flipped labels) captures how responsive a model is to the provided definition. \textbf{Rescue Rate} ($P(\text{correct}\mid \text{zero-shot wrong})$) quantifies how often prompting fixes errors. \textbf{Corruption Rate} ($P(\text{wrong}\mid \text{zero-shot correct})$) measures how often prompting breaks correct predictions. Models that are easily steered can be steered in both beneficial and harmful directions. Finally, \textbf{Prediction Bias} ($\text{predicted positive rate - true positive rate}$) captures whether prompting shifts predictions toward over-calling (more false positives) or under-calling (more false negatives) relative to the dataset's base rate. 

\vspace{-0pt}
\paragraph{Confidence.}
We elicit verbalized confidence scores by prompting the model to output a numerical score (0--100) alongside its prediction. The exact prompt suffix is shown in Figure~\ref{fig:confidence_prompt}. We use the same template across all models; only the label string varies by dataset. All calls use temperature~0.

\begin{figure}[t]
\centering
\small
\fbox{\parbox{0.92\linewidth}{%
\texttt{[Base classification prompt...]}\vspace{0.3em}\\
\texttt{Respond in exactly this format:}\\
\texttt{PREDICTION: [LABEL/not LABEL]}\\
\texttt{CONFIDENCE: [0-100]}
}}
\caption{Confidence elicitation prompt suffix appended to all classification prompts. \texttt{LABEL} is replaced with the dataset-specific positive class (e.g., \texttt{toxic}, \texttt{hateful}, \texttt{offensive}, etc.). The base prompt varies by condition (zero-shot, definition, few-shot, misaligned, etc.). This template is identical across all models.}
\label{fig:confidence_prompt}
\vspace{-6pt}
\end{figure}

\vspace{-0pt}
\subsection{Experimental Setup}
\label{sec:experimental_setup}

\vspace{-0pt}
\paragraph{Models.}
We evaluate nine instruction-tuned models spanning open-weights and
proprietary systems (Table~\ref{tab:models}), organized in two tiers.
Six models form the \emph{core} of our analysis, spanning dense
architectures (7B--70B parameters) and mixture-of-experts
architectures (Mixtral-8x7B, DeepSeek-V3) across the Llama and
Mistral families. All mixed-effects regressions (RQ2, RQ3) are scoped
to these six models to keep statistical inference consistent. We
additionally evaluate three \emph{extended} models---GPT-4o-mini,
Llama-3.3-70B, and Qwen-2.5-72B---under every prompting condition
on every primary dataset. These verify that our headline accuracy,
rescue-rate, and DSF findings extend to a broader, more recent slice
of proprietary and open-weights systems. For MoE models, we report
both total and active parameters per forward pass.
Appendix~\ref{app:coverage} details which models contribute to each
table and figure.

\vspace{-2pt}
\paragraph{Datasets.}
We focus on toxicity and hate speech detection across five primary
datasets covering a range of online communication environments, from
formal news comments to informal gaming chats, each with a different
operational definition of toxicity (Table~\ref{tab:datasets}). These
five form the analytic core and appear in every results table in the main paper. We additionally use three supplementary datasets for targeted robustness and generalization checks: \emph{Jigsaw Unintended
Bias}, which extends the DSF panel to $N=54$
model--dataset pairs and serves as a label-bias robustness check for the rescue-rate analysis; and \emph{SemEval-2018 Irony} 
together with \emph{Subjectivity}, which provide non-toxicity generalization checks for RQ1 and RQ2 (Appendix~\ref{app:generalization}). Appendix~\ref{app:coverage} maps
each dataset to the specific analyses it contributes to.

\begin{table}[t]
\caption{Five primary toxicity datasets followed by three supplementary datasets ($\dagger$). \% Pos. is the proportion of the task-positive label in the original dataset.}
\label{tab:datasets}
\vspace{-8pt}
\vskip 0.1in
\centering
\small
\begin{tabular}{@{}p{3.2cm}lrr@{}}
\toprule
\textbf{Dataset} & \textbf{Domain} & \textbf{Size} & \textbf{\% Pos.} \\
\midrule
Twitter Hate \newline {\scriptsize\cite{davidson2017automated}} & Social Media & 24,783 & 5.8 \\
OLID \newline {\scriptsize\cite{zampieri2019offenseval}} & Social Media & 14,100 & 33.2 \\
GameTox \newline {\scriptsize\cite{naseem2025gametox}} & Gaming & 53,000 & 42.5 \\
Fox News \newline {\scriptsize\cite{gao2017detecting}} & News & 1,528 & 28.5 \\
Jigsaw Toxic Comments \newline {\scriptsize\cite{jigsaw2018toxicity}} & Forum & 159,571 & 9.6 \\
\midrule
Jigsaw Unintended Bias$^\dagger$ \newline
{\scriptsize\cite{borkan2019nuanced}} & Forum & 97,320 & 8.0 \\
SemEval-2018 Irony$^\dagger$ \newline {\scriptsize\cite{vanhee2018semeval}} & Social Media & 4,618 & 48.1 \\
Subjectivity$^\dagger$ \newline {\scriptsize\cite{pang2004sentimental}} & Reviews & 10,000 & 50.0 \\
\bottomrule
\end{tabular}
\vskip -0.1in
\end{table}

\begin{table}[t]
\caption{Overview of models evaluated. Act.\ = parameters active per forward pass (MoE models). N/D = not disclosed. $^{*}$Estimated from unofficial sources; not officially disclosed by OpenAI.}
\label{tab:models}
\vspace{-8pt}
\vskip 0.1in
\centering
\small
\begin{tabular}{@{}llrr@{}}
\toprule
\textbf{Model} & \textbf{Arch.} & \textbf{Param.} & \textbf{Active} \\
\midrule
Llama-3.1-8B \newline {\scriptsize\cite{grattafiori2024llama3}} & Dense & 8B & 8B \\
Llama-3.1-70B \newline {\scriptsize~\cite{grattafiori2024llama3}} & Dense & 70B & 70B \\
Llama-3.3-70B \newline {\scriptsize\cite{meta2024llama33}} & Dense & 70B & 70B \\
Mistral-7B \newline {\scriptsize\cite{jiang2023mistral}} & Dense & 7B & 7B \\
Mistral-Small-24B \newline {\scriptsize~\cite{mistralai2025mistralsmall}} & Dense & 24B & 24B \\
Mixtral-8x7B \newline {\scriptsize\cite{jiang2024mixtral}} & MoE & 47B & 13B \\
DeepSeek-V3 \newline {\scriptsize\cite{deepseekv3}} & MoE & 671B & 37B \\
GPT-4o-mini \newline {\scriptsize\cite{openai2024gpt4omini}} & Prop. & 8B\rlap{$^{*}$} & 8B\rlap{$^{*}$} \\
Qwen-2.5-72B  {\scriptsize\cite{qwen2025qwen25}} & Dense & 72B & 72B \\
\bottomrule
\end{tabular}
\vskip -0.1in
\end{table}

\vspace{-2pt}
\paragraph{Experiments.}
We evaluate five prompting conditions: (1) Zero-Shot: task name only with no additional context, (2) Aligned Definition: task name plus the dataset's official definition of the target concept, (3) Few-Shot: task name plus 4 random balanced examples, (4) FS+Def: the aligned definition combined with 4 few-shot examples, and (5) Misaligned: definitions from other datasets applied to the target dataset (6 variants, detailed in Appendix~\ref{app:definitions}). All prompts follow a consistent template where the classification question is followed by the message and an ``Answer:'' suffix. We additionally run two DSPy~\citep{khattab2023dspy} automated optimization conditions on all nine models: DSPy Optimized (automatically selected few-shot examples) and DSPy Aligned (selected examples plus aligned definition). For each dataset-model pair, we randomly sampled 1,000 instances. All inference was conducted via API at temperature 0 for reproducibility.

\vspace{-2pt}
\paragraph{Statistical Analysis}
For RQ1, we examined associations between familiarity metrics and annotation accuracy using regression with cluster-robust standard errors (to account for dependence within models) and partial correlations controlling for dataset. For RQ2 and RQ3, we used mixed effects logistic regression on the six open-weights models to account for repeated observations of the same texts across conditions. This handles the binary outcome (correct vs. incorrect) while modeling the non-independence of the same text being evaluated under multiple prompting conditions. We include random intercepts for text and fixed effects for model, prompting condition, domain, and their interactions.

\vspace{-0pt}
\section{Results}
\vspace{-4pt}

Our core analysis covers predictions across 9 models, 5 toxicity datasets, and 10 prompting conditions, plus two DSPy automated optimization conditions. Table~\ref{tab:model_performance} summarizes model performance and steerability, while Table~\ref{tab:glm_results} presents key findings from our mixed effects regression models.

\begin{table*}[t]
\caption{Model performance (Accuracy\,\%\,/\,AUC) across prompting conditions. Misaligned averages 6 definition-swap conditions (full breakdown in Appendix~\ref{app:performance}).}
\label{tab:model_performance}
\vspace{-8pt}
\vskip 0.1in
\centering
\setlength{\tabcolsep}{3pt}
\footnotesize
\begin{tabular}{@{}l|cccccc|c|c@{}}
\toprule
\textbf{Model} & \textbf{Zero-Shot} & \textbf{Aligned Def} & \textbf{Few-Shot} & \textbf{FS + Def} & \textbf{DSPy Opt.} & \textbf{DSPy Aligned} & \textbf{Misaligned} & \textbf{Overall} \\
\midrule
Llama-3.1-8B      & 76.7/.832 & 80.1/.843 & 75.3/.822 & 75.4/.824          & $72.3^{\dagger}$/.809 & 73.5/.821 & 77.0/.822 & 76.8/.818 \\
Llama-3.1-70B     & 79.8/.881 & 82.1/.888 & 81.2/.892 & 82.1/.898          & 79.9\phantom{$^{\dagger}$}/\textbf{.889} & 82.6/.899 & 78.1/.859 & 79.7/.869 \\
Llama-3.3-70B     & 80.5/\textbf{.885} & 82.3/\textbf{.892} & 82.0/\textbf{.896} & 83.0/\textbf{.904} & $77.5^{\dagger}$/.884 & 81.3/\textbf{.904} & 79.0/.863 & 80.1/\textbf{.874} \\
Mistral-7B        & 80.5/.858 & 81.3/.787 & 81.5/.852 & 80.8/.838          & 82.0\phantom{$^{\dagger}$}/.827 & 81.7/.825 & \textbf{81.4}/.823 & 81.2/.819 \\
Mistral-Small-24B & 78.0/.868 & 81.0/.863 & 80.8/.870 & 82.3/.879          & 79.3\phantom{$^{\dagger}$}/.867 & 81.2/.876 & 80.7/.850 & 80.5/.852 \\
Mixtral-8x7B      & 81.0/.868 & 82.7/.868 & 82.4/.868 & 80.9/.859          & 82.8\phantom{$^{\dagger}$}/.860 & 82.7/.861 & 81.2/.843 & 81.6/.850 \\
DeepSeek-V3       & 81.3/.874 & 83.0/.881 & 82.6/.890 & \textbf{83.8}/.893 & 80.9\phantom{$^{\dagger}$}/.877 & 83.3/.890 & 80.7/.856 & 81.6/.865 \\
GPT-4o-mini       & 81.6/.871 & \textbf{83.3}/.875 & \textbf{84.1}/.880 & 83.3/.880 & \textbf{84.4}\phantom{$^{\dagger}$}/.883 & \textbf{84.3}/.883 & 81.1/.857 & \textbf{82.3}/.862 \\
Qwen-2.5-72B      & \textbf{83.3}/.876 & 82.2/.879 & 83.8/.894 & 83.2/.892 & 82.5\phantom{$^{\dagger}$}/.887 & 83.2/.891 & 81.2/\textbf{.868} & 82.2/.872 \\
\midrule
\textbf{Cond. Avg} & 80.3/.868 & 82.0/.864 & 81.5/.874 & 81.6/.874 & 80.2\phantom{$^{\dagger}$}/.865 & 81.5/.872 & 80.0/.849 & 80.8/.853 \\
\bottomrule
\end{tabular}
\\[3pt]
{\scriptsize $^{\dagger}$DSPy Opt.\ degrades Llama-3.1-8B (76.7$\to$72.3) and Llama-3.3-70B (80.5$\to$77.5), consistent with~\citet{tan2025langprobe}.}
\vskip -0.1in
\end{table*}

\vspace{-0pt}
\subsection{Answering RQ1: Impact of Task Familiarity}
\label{sec:rq1}
\vspace{-4pt}

We analyzed whether annotation performance correlates more strongly with text memorization or semantic task-definition alignment (Definition-Specific Familiarity, DSF). The unit of analysis is a model--dataset cell: text-reproduction scores are first averaged over examples within each cell, just as accuracy is, while DSF is measured for that model and definition. We then partial out dataset identity. The resulting coefficient therefore answers a within-task question: after removing average differences among datasets, do models whose elicited concept is closer to a dataset's definition perform better on that same task?

In zero-shot evaluation, consensus DSF is positively associated with accuracy (partial $r=+0.41$, 95\% CI $[+0.15,+0.62]$, $p=0.003$, $N=54$ model--dataset cells). This is not an artifact of one sentence encoder: the six encoder-specific estimates are MiniLM $+.36$, MPNet $+.33$, BGE-large $+.49$, E5-large $+.43$, Instructor-large $+.30$, and OpenAI $+.43$ (Appendix~\ref{app:dsf_embeddings}). Their DSF score vectors are also highly concordant (pairwise $r=.88$--$.99$), showing that the encoders largely agree about which model--definition pairs are aligned.

\begin{figure}[t]
\centering
\includegraphics[width=\columnwidth]{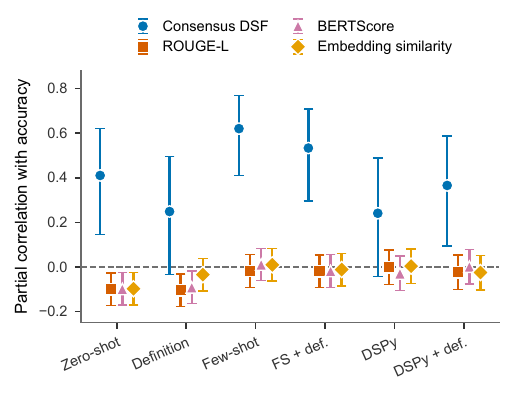}
\vspace{-20pt}
\caption{Partial correlations controlling for dataset. DSF uses cell-level accuracy ($N=54$ per condition); memorization metrics use per-text correctness ($N=722$; DSPy: $N=641$). Bars show 95\% CIs.}
\label{fig:dsf_conditions}
\vspace{-6pt}
\end{figure}

Figure~\ref{fig:dsf_conditions} extends this comparison across six prompting conditions. DSF is strongest with few-shot examples (partial $r=+0.62$, 95\% CI $[+0.41,+0.77]$) and few-shot examples plus the definition ($r=+0.53$, $[+0.30,+0.71]$). It remains positive in the other conditions: DSPy prompts ($r=+0.24$ without and $+0.37$ with the definition), zero-shot, and definition-only prompting. The three per-text memorization measures show a different pattern: after controlling for dataset, their correlations with whether an individual text is classified correctly are small and generally non-positive across conditions. The consistently positive DSF correlations show that DSF captures meaningful within-task differences between models, identifying which models are better suited to a given task rather than merely reflecting dataset difficulty irrespective of prompting technique used.

Table~\ref{tab:familiarity_correlations} contrasts this conceptual-alignment signal with three text-reproduction measures in the zero-shot condition. Their strong raw correlations largely reflect between-dataset differences: Jigsaw, for example, is both relatively high-memorization and difficult. Once dataset identity is controlled for, ROUGE-L, BERTScore, and embedding similarity show no positive association with performance (partial $r=-.19,-.15,-.16$, respectively), whereas consensus DSF remains positive. \emph{\textbf{Models whose elicited concepts more closely match a dataset's definition tend to perform better on that task, while text memorization does not provide the same signal.}}

The same positive association appears in supplementary experiments on irony and subjectivity detection (partial $r=+0.34$, $N=18$ model--dataset pairs), providing directional evidence that the DSF result is not confined to toxicity; see Appendix~\ref{app:generalization}.

\begin{table}[th!]
\caption{Correlations between familiarity metrics and zero-shot annotation accuracy on the extended $N=54$ panel. Raw correlations are misleading due to dataset confounds; partial correlations (controlling for dataset) reveal that definition alignment (consensus DSF) is positively associated with performance while three distinct memorization metrics are not.} %BERTScore and embedding similarity use the same all-MiniLM-L6-v2 encoder as DSF, so any sign difference reflects signal type rather than encoder choice.
\label{tab:familiarity_correlations}
\vspace{-8pt}
\vskip 0.1in
\centering
\small
\begin{tabular}{@{}lcc@{}}
\toprule
\textbf{Metric} & \textbf{Raw $r$} & \textbf{Partial $r$} \\
\midrule
Text Memorization (ROUGE-L)            & $-0.80$ & $-0.19$ \\
Memorization (BERTScore)               & $-0.76$ & $-0.15$ \\
Memorization (Embedding Similarity)    & $-0.71$ & $-0.16$ \\
Definition Familiarity (consensus DSF) & $+0.74$ & $+0.41$ \\
\bottomrule
\end{tabular}
\vskip -0.1in
\end{table}

\subsection{Answering RQ2: Impact of Knowledge Injection}
\label{sec:rq2}
\vspace{-4pt}

Table~\ref{tab:glm_results} presents the factors associated with both annotation correctness and error correction, evaluated via mixed-effects logistic regression. Domain is most strongly associated with correctness: compared with Forum, the odds are approximately $14\times$ higher for Gaming ($OR=13.85$) and $5\times$ higher for Social Media ($OR=5.45$). Aligned definitions improve mean accuracy by only 2.2 points over zero-shot, and DSPy changes the corresponding averages by at most 0.5 points (Table~\ref{tab:model_performance}). Moreover, DSPy degrades both Llama models, consistent with prior findings that prompt optimizers can underperform unoptimized baselines~\citep{tan2025langprobe}. These modest, heterogeneous gains suggest that suboptimal prompt design alone does not explain the performance ceiling. Conditional on other covariates, a correct zero-shot response predicts $6.4\times$ higher odds of correctness with definitions or examples ($OR=6.43$), suggesting that prompting consolidates correct answers more readily than it rescues errors.

\vspace{-2pt}
\begin{table}[th!]
\caption{Key findings from mixed effects logistic regression. OR $>$ 1 indicates higher odds of a correct annotation. Reference categories: Forum (domain), Llama-8B (model). Full model specifications in Appendix~\ref{app:regression}. All $p < 0.001$.}
\label{tab:glm_results}
\vspace{-8pt}
\vskip 0.1in
\centering
\small
\begin{tabular}{@{}l d{2}@{\hspace{4pt}}l@{}}
\toprule
\textbf{Factor} & \multicolumn{2}{c}{\textbf{OR [95\% CI]}} \\
\midrule
\multicolumn{3}{@{}l}{\textit{Factors Associated with Correctness:}} \\
\quad Gaming (vs. Forum) & 13.85 & [13.31, 14.41] \\
\quad Social Media (vs. Forum) & 5.45 & [5.27, 5.63] \\
\quad Aligned Definition & 1.08 & [1.04, 1.11] \\
\quad Aligned $\times$ News & 1.91 & [1.80, 2.02] \\
\quad \textbf{Zero-shot Correct} & 6.43 & [6.28, 6.58] \\
\midrule
\multicolumn{3}{@{}l}{\textit{Factors Associated with Error Correction:}} \\
\quad ZS Confidence (per SD) & 0.84 & [0.82, 0.87] \\
\bottomrule
\end{tabular}
\vskip -0.1in
\end{table}

\begin{table}[ht!]
\caption{Rescue rate by model, aggregated across prompted conditions. Rescue rate is $P(\text{correct} \mid \text{prompted}, \text{zero-shot wrong})$.}
\label{tab:rescue_by_model}
\vspace{-8pt}
\vskip 0.1in
\centering
\small
\begin{tabular}{@{}lccc@{}}
\toprule
\textbf{Model} & \textbf{Rescue Rate} & \textbf{95\% CI} & \textbf{N (ZS err.)} \\
\midrule
Mistral-small-24B & 44.2\% & [43.1, 45.2] & 8,536 \\
DeepSeek-V3 & 38.1\% & [36.9, 39.2] & 7,264 \\
Llama-3.1-8B & 37.5\% & [36.5, 38.5] & 9,040 \\
GPT-4o-mini & 35.7\% & [34.6, 36.8] & 7,152 \\
Mixtral-8x7B & 34.6\% & [33.5, 35.7] & 7,416 \\
Mistral-7B & 31.6\% & [30.5, 32.6] & 7,568 \\
Llama-3.1-70B & 31.3\% & [30.3, 32.4] & 7,848 \\
Llama-3.3-70B & 30.4\% & [29.4, 31.4] & 7,592 \\
Qwen-2.5-72b & 27.8\% & [26.8, 28.9] & 6,504 \\
\bottomrule
\end{tabular}
\vskip -0.1in
\end{table}

\begin{figure}[h]
    \centering
    \includegraphics[width=\linewidth]{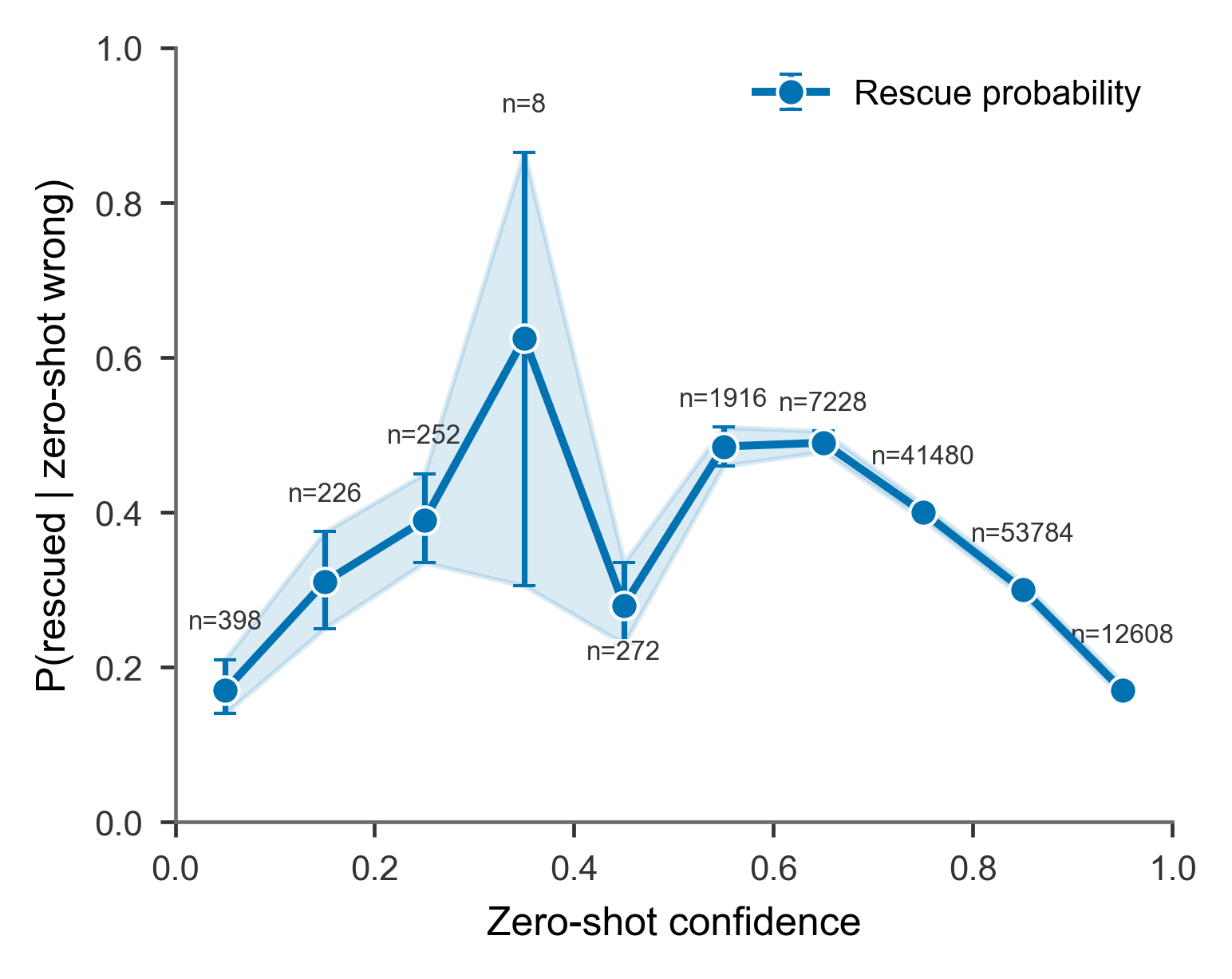}
    \vspace{-2.0em}
    \caption{Rescue probability vs.\ zero-shot confidence for zero-shot errors. The inverted-U shape exhibits two distinct failure modes (analyzed in \hyperref[sec:confidence_stickiness]{\emph{Confidence and Decision Stickiness}}): \emph{decision stickiness} at high confidence (right tail) and an out-of-distribution effect at very low confidence (left tail).}
    \label{fig:rescue_confidence}
    \vspace{-8pt}
\end{figure}

\phantomsection
\label{sec:confidence_stickiness}
\emph{Rescue Rate} is the probability that prompting corrects a zero-shot error. The overall rescue rate is only 34.8\%, meaning nearly two-thirds of errors resist correction. The same qualitative pattern appears on irony and subjectivity: few-shot prompting rescues 45.0\% of $3{,}052$ zero-shot errors, leaving most errors uncorrected (Appendix~\ref{app:generalization}).
Figure~\ref{fig:rescue_confidence} shows an inverted-U: rescue probability peaks at 51.8\% for moderate confidence (0.6--0.7) and falls sharply to 20.8\% for high-confidence errors ($>0.9$), the most statistically reliable region of the curve. Model 4 corroborates this trend: each standard-deviation increase in zero-shot confidence reduces rescue odds by 16\% ($OR=0.84$, Table~\ref{tab:glm_results}). \emph{\textbf{When models are confidently wrong, prompting rarely helps.}}

The drop at the extreme low-confidence end ($n<$1{,}000) is a distinct failure mode rather than a counterexample to decision stickiness. These instances are out-of-distribution and context-deficient: their ROUGE-L is far below the rest of the data (0.025 vs.\ 0.061) and they are dramatically shorter (median 6 vs.\ 22 words), consisting largely of single-word gaming fragments (e.g., ``noob'', ``wtf'', ``bot'') whose toxicity depends on context that is unavailable to the model.

Importantly, steerability is not a monotonic function of model size: for example, Mistral-Small-24B rescues more zero-shot errors than Llama-3.1-70B in our experiments, despite being substantially smaller (Table~\ref{tab:rescue_by_model}). To confirm this finding is not an artifact of label-bias issues known to affect the standard Jigsaw dataset (e.g., systematic flagging of African American Vernacular English as toxic), we replicated the rescue rate analysis on the Jigsaw Unintended Bias dataset~\citep{borkan2019nuanced}, which was explicitly designed to reduce identity-group annotation bias. Rescue rates remain uniformly below 50\% across all nine models (Table~\ref{tab:jigsaw_rescue}), confirming that decision stickiness is not specific to Jigsaw's label properties.

\subsection{Answering RQ3: Misalignment Impact}
\label{sec:rq3}

We analyze 6 misaligned definition conditions to investigate how LLMs respond to incorrect instructions (Table~\ref{tab:misalignment_results}). The results reveal several surprising patterns.

We find that LLMs are responsive to definition scope. Narrow definitions (requiring targeting based on race, religion, gender, etc.) produce prediction bias of $-7\%$ to $-12\%$ (under-prediction), while broad definitions produce bias of $+9\%$ to $+13\%$ (over-prediction). This shift in prediction rates demonstrates that LLMs do not simply ignore misaligned instructions; they adjust decision threshold to match the provided criteria. This responsiveness is a double-edged sword: while it enables definition-based steering, it also means models will confidently apply incorrect definitions.

The worst misaligned condition (gaming toxicity: 76.4\%) is 5.6\% below aligned, while the best (Fox News hate speech: 82.6\%) actually \emph{outperforms} the aligned definition by 0.6\%. Knowledge built up during training and tuning constrains how much definitions can shift behavior: even with ``wrong'' definitions, models remain relatively stable, indicating persistent influence from these internalized task concepts.

Surprisingly, some misaligned definitions provide net benefit over zero-shot (rescuing more errors than they corrupt). For instance, using a ``hate speech'' definition can improve performance on general toxicity tasks by focusing the model on clearer criteria, even if technically misaligned. This finding challenges the assumption that the ``correct'' definition is always optimal - empirically, some misaligned definitions better match the model's internal concept.

\begin{table}[th!]
\caption{Misalignment analysis: LLMs adjust prediction thresholds based on definition, yet maintain high confidence. Rows are ordered by definition breadth (Narrow $\to$ Broad). Confidence remains 84--91\% regardless of alignment. All values are percentages.}
\label{tab:misalignment_results}
\vspace{-8pt}
\vskip 0.1in
\centering
\small
\sisetup{detect-weight=true, detect-family=true}
\begin{tabular}{@{}l
  S[table-format=2.1]
  S[table-format=2.1]
  S[table-format=+2.1, print-implicit-plus]
  S[table-format=2.1]@{}}
\toprule
{\textbf{Condition}} & {\textbf{Acc.}} & {\textbf{Change}} & {\textbf{Bias}} & {\textbf{Conf.}} \\
\midrule
Aligned definition & 82.0 & 11.3 & {ref} & 88.1 \\
\midrule
Fox News hate speech & \bfseries 82.6 & 17.5 & -7.2 & \bfseries 90.6 \\
Twitter hate speech & 78.5 & \bfseries 23.1 & \bfseries -12.1 & 88.1 \\
GameTox toxicity & 79.6 & 15.4 & +0.5 & 86.5 \\
Olid offensive & 81.9 & 10.9 & +5.1 & 89.1 \\
General toxicity & 81.1 & 9.2 & +8.8 & 85.0 \\
Gaming toxicity & 76.4 & 13.3 & \bfseries +13.1 & 88.5 \\
\bottomrule
\end{tabular}
\vskip -0.1in
\end{table}

\begin{figure}[h]
    \centering
    \includegraphics[width=\linewidth]{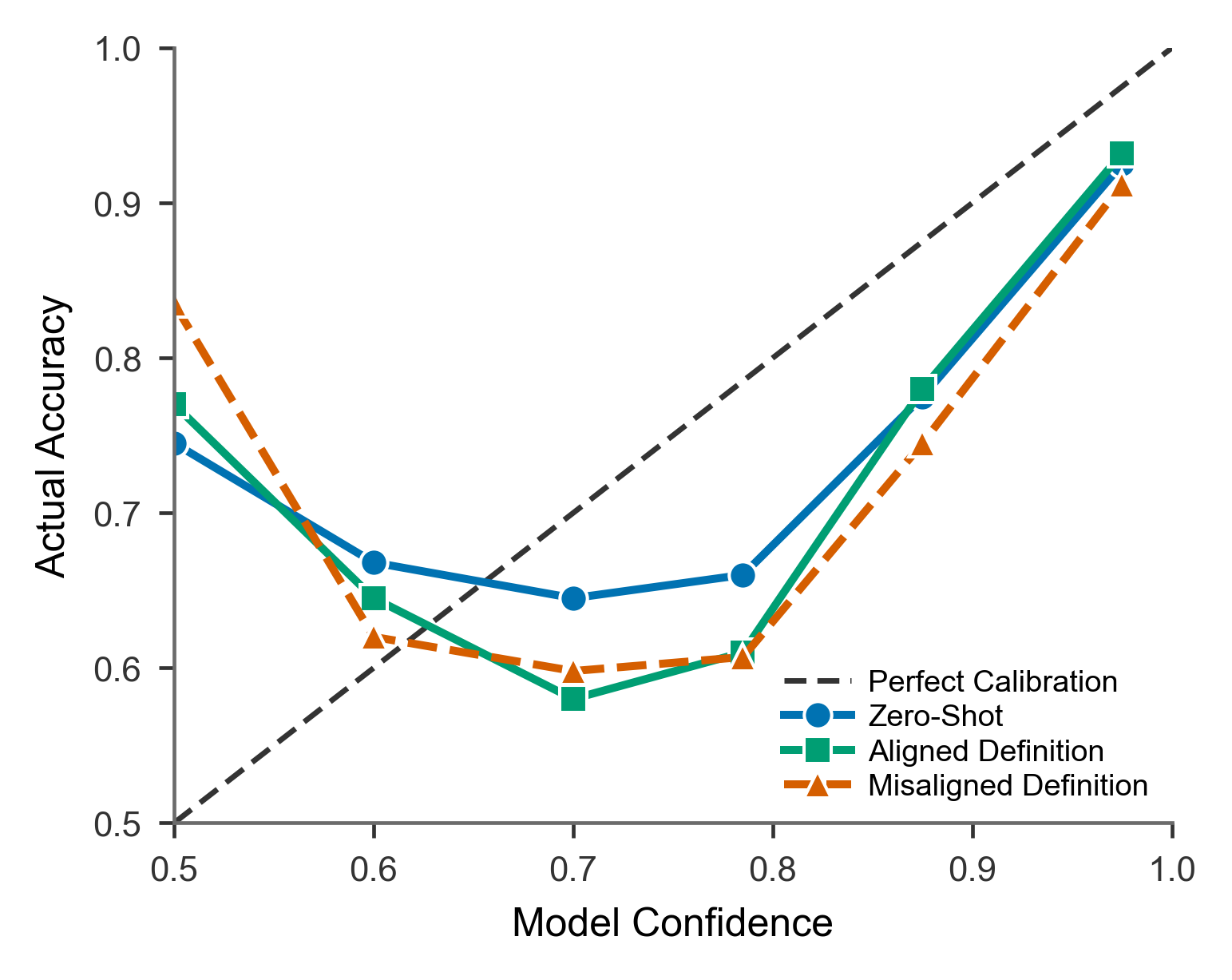}
    \vspace{-2.0em}
    \caption{Calibration curves showing confidence vs.\ actual accuracy across conditions. All conditions exhibit overconfidence (below the diagonal), with no meaningful separation between aligned and misaligned definitions: models cannot distinguish when they are applying incorrect instructions.}
    \label{fig:calibration_curve}
    \vspace{-10pt}
\end{figure}

\textbf{Critical Calibration Failure.} Models \emph{\textbf{remain confident even when applying misaligned definitions}} (Table~\ref{tab:misalignment_results}): mean confidence changes little from zero-shot (87.0\%) across aligned and misaligned definitions (85.0--90.6\%). The narrowest definition (Fox News hate speech) produces the \emph{highest} confidence, not the lowest. This represents a calibration failure: models should exhibit increased uncertainty when instructions contradict their training distribution, but they do not. As shown in Figure~\ref{fig:calibration_curve}, all conditions (zero-shot, aligned, and misaligned) exhibit similar overconfidence patterns, with calibration curves falling below the diagonal. Critically, there is no meaningful separation between aligned and misaligned definitions, meaning \emph{\textbf{practitioners cannot rely on confidence scores to detect definition errors}}.

Across all models and datasets, definition choice produces 17\% accuracy variation (e.g., for Fox News, accuracy ranges from 61.6\% with the ``Gaming Toxicity'' definition to 78.4\% with the ``Twitter Hate Speech'' definition), while model choice produces only 5\% variation on average. This implies that for annotation tasks, \emph{\textbf{careful definition design is more impactful than selecting larger or more capable models}}.

Models showing high rescue rates under misalignment also show high corruption rates (Table~\ref{tab:rescue_by_model} and Table~\ref{tab:corruption_matrix}). Mistral-Small-24B achieves the highest rescue rate (73.7\% with Twitter hate speech) but also the highest corruption rate (21.0\%). This creates a fundamental trade-off: \emph{\textbf{models that are easily steered by definitions can be steered in both beneficial and harmful directions}}.

\section{Discussion}
\label{sec:discussion}
\vspace{-4pt}

DSF predicts accuracy after accounting for dataset differences, whereas three text-memorization metrics do not. This supports a concept-level account of annotation performance and shifts validation from auditing text overlap toward assessing whether a model's concept boundary matches the intended definition.

\vspace{-0.5em}
\paragraph{Prompting has limited ability to overturn wrong decisions.}
We find substantial \emph{decision stickiness}: most zero-shot errors persist under aligned definitions and few-shot examples, and high-confidence errors are especially resistant to correction (Figure~\ref{fig:rescue_confidence}, Table~\ref{tab:glm_results}). This indicates that prompting primarily stabilizes or modestly improves predictions the model already gets right, rather than serving as a reliable mechanism for error recovery. We further test whether iterative, history-aware prompting can break this \emph{stickiness}: a three-turn rescue sequence that progressively adds few-shot examples, the aligned definition, and an explicit reconsideration prompt raises rescue from 7.5\% at Turn 1 to only 18.7\% at Turn 3, and lifts high-confidence rescue to just 8.5\% (Appendix~\ref{app:multiturn}). Decision stickiness therefore persists under iterative correction, not only one-shot prompting. Notably, steerability is not monotonic in model size (Table~\ref{tab:rescue_by_model}), implying that larger models are not necessarily easier to steer for annotation workloads.

\vspace{-0.5em}
\paragraph{Misaligned definitions shift behavior without reliably signaling risk.}
When given semantically related but mismatched definitions, models change their prediction rates in systematic ways (Table~\ref{tab:misalignment_results}, see also Table~\ref{tab:dataset_misaligned_acc} in Appendix), consistent with literal application of the provided criteria. However, accuracy often degrades only modestly and sometimes even improves relative to the aligned condition. One interpretation is that the model's training supplies a strong prior that anchors behavior, providing a performance floor even under imperfect instructions. Another possibility is that dataset label boundaries may be closer to some alternative formulations than the dataset's stated definition, especially for inherently subjective constructs. Regardless, the key operational point is that \emph{definition wording is a first-order experimental factor}: in our study, definition choice can induce substantial performance variation, so treating the definition as fixed can meaningfully understate uncertainty in downstream analyses.

\vspace{-0.5em}
\paragraph{Confidence is conditional, not diagnostic of definition validity.}
A central deployment risk is that models remain highly confident under misalignment, and calibration curves show little separation between aligned and misaligned conditions (Figure~\ref{fig:calibration_curve}). Importantly, this does \emph{not} imply that the model ``should know'' the definition is wrong; rather, it suggests that the reported confidence reflects certainty \emph{given the provided instructions}, not a meta-uncertainty about whether those instructions match the intended labeling policy. As a result, confidence thresholds are poorly suited for detecting definition errors or definition-dataset mismatch.

\vspace{-0.5em}
\paragraph{Practical implications for LLM annotation pipelines.}
These findings suggest three lightweight safeguards. First, measure \emph{definition alignment} before large-scale annotation (e.g., DSF-style checks) to anticipate which model--definition pairs are likely to work well. Second, \emph{stress-test definitions} by evaluating multiple plausible formulations and reporting sensitivity (e.g., changes in bias, rescue, and corruption), rather than assuming a single ``correct'' wording is robust. Third, avoid using confidence as a proxy for definition appropriateness.

\vspace{-0.5em}
\paragraph{Limitations and future work.}
Our study focuses primarily on binary toxicity classification with concept-substitution misalignment. Supplementary irony and subjectivity analyses extend beyond toxicity, but multi-class, span-labeling, and open-ended tasks may have different failure modes. Consensus DSF across six encoders reduces embedding-choice sensitivity (Appendix~\ref{app:dsf_embeddings}), but alternative similarity measures and elicitation prompts need further exploration. The predictive validity of LLM self-reports also depends on instrument specificity and conversational context \citep{kocielnik2026rethinking}, motivating tests of whether DSF remains informative across elicitation contexts. Because all models are instruction-tuned, we cannot separate capability limits from intentional safety constraints or identify which training stage produces concept anchoring. Addressing this requires matched base and post-trained models.
Domain-adaptive pretraining has improved context-sensitive classification \citep{schurgerfoy2026context}; whether such adaptation also improves definition fidelity remains an open question. Results measure fidelity to dataset-specific annotation policies rather than normative correctness, although replication on Jigsaw Unintended Bias and stability across datasets reduce dependence on any single policy's ambiguity or bias. Finally, our evidence is correlational: causal claims would require interventions that manipulate DSF while holding other factors fixed. Future work should test stronger forms of misalignment, alternative multi-turn correction strategies such as self-consistency or critique-and-revise, and scalable procedures for selecting model--definition pairs under explicit constraints, all of which present significant trade-offs in deployment cost and complexity even if proven effective.

\vspace{-2pt}
\section{Conclusion}
\vspace{-2pt}

We study how model-internalized priors interact with prompt-based steering in LLM annotation. Across 9 models and 5 toxicity datasets, we find that performance is best explained by \emph{definition-specific familiarity} (DSF): controlling for dataset difficulty, DSF is positively associated with accuracy (partial $r=+0.41$, $N=54$), while three distinct memorization metrics (ROUGE-L, BERTScore, embedding similarity) are not. Prompting often fails to correct mistakes, as most zero-shot errors persist even under aligned definitions and few-shot examples. Models can confidently follow \emph{misaligned} definitions, making confidence an unreliable indicator of definition correctness. Finally, definition wording induces larger performance swings than model choice, highlighting definition design as crucial. Practically, LLM annotation pipelines should prioritize measuring definition alignment as an early indicator of performance rather than relying on larger models or confidence-based strategies.

\vspace{-5pt}
\section*{Acknowledgment}
\vspace{-3pt}
This work is supported by the Caltech Linde Center for Science, Society, and Public Policy (LCSSP). R. Michael Alvarez is Flintridge Foundation Professor of Political and Computational Social Science at Caltech. We thank Andrea Boonyarungsrit, Grant Cahill, Jonathan Lane, Min Kim, and Fereshteh Soltani for prior collaboration that helped inform some of the motivating challenges behind this work. The views and opinions expressed in this paper are solely those of the authors and do not necessarily reflect those of the individuals acknowledged or their affiliated organizations.

%The motivation for this project was partly informed by challenges encountered in prior work arising from a Caltech--Activision collaboration, which is cited separately in this paper.

\vspace{-5pt}
\section*{Impact Statement}
\vspace{-3pt}
This work examines how model-internalized priors interact with prompt-based steering in LLM annotation tasks. Our findings have implications for practitioners deploying LLMs as annotators or judges. Positively, identifying failure modes such as decision stickiness and miscalibration under definition misalignment can help practitioners develop better validation procedures and avoid over-reliance on confidence scores. However, our results also reveal that models will confidently apply incorrect definitions when provided, which could lead to systematic errors if task specifications are poorly designed.

\bibliographystyle{icml2026}
\bibliography{references}

\newpage
\appendix
\onecolumn

\section{Definition Texts Used}
\label{app:definitions}

This section provides the exact definition texts given to models, organized by scope (narrow to broad). Understanding definition scope is critical for interpreting misalignment results.

\subsection{Narrow Scope Definitions}
These definitions require identity-based targeting; general insults without identity targeting are explicitly excluded.

\paragraph{Twitter Hate Speech \citep{davidson2017automated}.}
\begin{quote}
\textbf{Hate Speech:} Language that expresses hatred towards a targeted group or is intended to be derogatory, to humiliate, or to insult the members of the group based on attributes such as race, religion, ethnic origin, sexual orientation, disability, or gender.

\textbf{Not Hate Speech:} Content that does not express hatred or target groups based on identity attributes. Note: Offensive language that contains profanity or insults but does not target identity groups is NOT considered hate speech.
\end{quote}

\paragraph{Fox News Hate Speech.}
\begin{quote}
\textbf{Hateful:} Language that explicitly or implicitly threatens or demeans a person or a group based upon a facet of their identity such as gender, ethnicity, or sexual orientation. This includes comments that express prejudice, discrimination, or hostility toward individuals or groups based on their protected characteristics.

\textbf{Non-hateful:} Comments that do not contain language that threatens or demeans individuals or groups based on their identity characteristics.
\end{quote}

\subsection{Medium Scope Definitions}
These definitions include insults and threats regardless of identity targeting, but still require explicit negative content.

\paragraph{GameTox Multi-Category Definition.}
\begin{quote}
\textbf{Toxic:} Any message that falls into one or more of these categories:
(1) Hate and Harassment: Utterances with identity-based hate or harassment;
(2) Threats: Utterances with threats of violence, physical harm, terrorism, or doxing;
(3) Extremism: Utterances with extremist views or recruitment attempts;
(4) Insults and Flaming: Insults or attacks on another player (not identity-based);
(5) Other Offensive Texts: Any message that harms reasonable enjoyment of the game.

\textbf{Not toxic:} Utterances without any toxicity from the above categories.
\end{quote}

\paragraph{OLID Offensive Language \citep{zampieri2019offenseval}.}
\begin{quote}
\textbf{Offensive:} This post contains offensive language or a targeted (veiled or direct) offense. This includes any form of non-acceptable language (profanity), insults, threats, and posts containing profane language or swear words.

\textbf{Not Offensive:} This post does not contain offense or profanity. The content may express opinions or criticism but does not include any form of offensive language, insults, or threats.
\end{quote}

\subsection{Broad Scope Definitions}
These definitions capture any content that is ``rude,'' ``disrespectful,'' or ``discouraging'' - very broad criteria.

\paragraph{Gaming Toxicity (CONDA) \citep{weld2021conda}.}
\begin{quote}
\textbf{Toxic:} Utterances containing explicit or implicit toxicity in gaming context. \emph{Explicit Toxicity:} Direct toxic expressions such as profanity, personal attacks, or insults directed at other players. \emph{Implicit Toxicity:} Subtler forms including sarcasm, veiled insults, or coded language that requires conversational context to identify as harmful.

\textbf{Not Toxic:} Utterances related to gameplay actions, neutral communication, or content that does not discourage player engagement. General expressions of frustration not directed at specific players (e.g., ``damn!'' or ``ugh'') are not toxic.
\end{quote}

\paragraph{General Toxicity (Perspective API/Jigsaw) \citep{jigsaw2018toxicity}.}
\begin{quote}
A rude, disrespectful, or unreasonable comment that is likely to make someone leave a discussion.
\end{quote}

\begin{table}[H]
\caption{Definition scope classification. \textbf{Narrow} definitions require identity-based targeting (e.g., race, religion, gender) to qualify as toxic, excluding general insults. \textbf{Medium} definitions include direct insults, threats, and profanity regardless of identity targeting. \textbf{Broad} definitions capture any content that is rude, disrespectful, or discouraging, including implicit and contextual toxicity. When applied to general toxicity tasks, narrow definitions produce under-prediction while broad definitions produce over-prediction.}
\label{tab:definition_scope}
\centering
\small
\begin{tabular}{@{}ll@{}}
\toprule
\textbf{Definition} & \textbf{Scope} \\
\midrule
Twitter Hate Speech & Narrow \\
Fox News Hate Speech & Narrow \\
GameTox Toxicity & Medium \\
OLID Offensive & Medium \\
Gaming Toxicity (CONDA) & Broad \\
General Toxicity & Broad \\
\bottomrule
\end{tabular}
\end{table}

\subsection{Non-Safety Task Definitions}
\label{app:definitions_nonsafety}
The following definitions were used in the generalization experiments reported in Appendix~\ref{app:generalization}. The irony and subjectivity definitions were used as the ``aligned definition'' prompt in the annotation experiments (Tables~\ref{tab:nonsafety_accuracy} and~\ref{tab:nonsafety_rescue}); the five stance definitions were used as the elicitation context for DSF on stance detection. All definitions operationalize the target concept following the framing of the original task papers and do not fit the narrow/medium/broad scope schema of Table~\ref{tab:definition_scope}, which applies only to the toxicity datasets.

\paragraph{Irony Detection (SemEval-2018 Task 3) \citep{vanhee2018semeval}.}
\begin{quote}
\textbf{Ironic:} A statement where the intended meaning is opposite to or incongruent with the literal meaning, often used to express humor, criticism, or sarcasm. Irony involves a discrepancy between what is said and what is meant, or between expectations and reality.

\textbf{Not Ironic:} A statement where the intended meaning matches the literal meaning, with no incongruence between expression and intent.
\end{quote}

\paragraph{Subjectivity Detection \citep{pang2004sentimental}.}
\begin{quote}
\textbf{Subjective:} A sentence that expresses personal opinions, evaluations, emotions, or speculations. Subjective sentences reflect the author's point of view, feelings, or interpretations rather than objective facts.

\textbf{Objective:} A sentence that presents factual information that can be verified independently of the author's personal opinion or emotional state.
\end{quote}

\paragraph{Stance Detection (SemEval-2016 Task 6) \citep{mohammad2016semeval}.}
The stance task uses five targets, each with its own definition. Unlike the other tasks in our study, stance is framed as a choice between two explicit alternatives (\emph{in favor} / \emph{against}) rather than as classification against a single concept definition, as discussed in Appendix~\ref{app:generalization}.

\begin{quote}
\textbf{Stance toward Legalization of Abortion.} \emph{In favor:} The tweet expresses support for or agreement with the legalization of abortion, including women's right to choose. \emph{Against:} The tweet expresses opposition to the legalization of abortion, including pro-life positions or moral objections.
\end{quote}

\begin{quote}
\textbf{Stance toward Atheism.} \emph{In favor:} The tweet expresses support for, agreement with, or positive views toward atheism or non-belief in God. \emph{Against:} The tweet expresses opposition to, criticism of, or negative views toward atheism.
\end{quote}

\begin{quote}
\textbf{Stance toward Climate Change.} \emph{In favor:} The tweet expresses belief in or concern about climate change as a real and serious issue requiring action. \emph{Against:} The tweet expresses skepticism about, denial of, or opposition to claims about climate change.
\end{quote}

\begin{quote}
\textbf{Stance toward Feminist Movement.} \emph{In favor:} The tweet expresses support for or agreement with the feminist movement and gender equality. \emph{Against:} The tweet expresses opposition to, criticism of, or negative views toward the feminist movement.
\end{quote}

\begin{quote}
\textbf{Stance toward Hillary Clinton.} \emph{In favor:} The tweet expresses support for, agreement with, or positive views toward Hillary Clinton. \emph{Against:} The tweet expresses opposition to, criticism of, or negative views toward Hillary Clinton.
\end{quote}

\section{Prompts}
\label{app:prompts}

All prompts follow a consistent template structure. The base classification prompt presents the task, followed by the text to classify, and ends with an ``Answer:'' suffix. The confidence elicitation suffix (Figure~\ref{fig:confidence_prompt}) is appended to all classification prompts.

\paragraph{Experimental Conditions.}
Table~\ref{tab:conditions} summarizes the 10 manually-specified prompting conditions together with the 2 DSPy automated optimization conditions used in our experiments.

\begin{table}[H]
\caption{Description of experimental conditions.}
\label{tab:conditions}
\centering
\small
\begin{tabular}{@{}llc@{}}
\toprule
\textbf{Condition} & \textbf{Description} & \textbf{Aligned} \\
\midrule
zero\_shot & No definition provided & -- \\
aligned\_definition & Definition matches dataset & Yes \\
few\_shot & 4 examples, no definition & -- \\
few\_shot\_aligned\_def & 4 examples + aligned definition & Yes \\
misaligned\_gaming\_toxicity & Gaming toxicity definition & No \\
misaligned\_general\_toxicity & Perspective API definition & No \\
misaligned\_gametox\_toxicity & GameTox definition & No \\
misaligned\_foxnews\_hate\_speech & Fox News hate speech definition & No \\
misaligned\_twitter\_hate\_speech & Twitter hate speech definition & No \\
misaligned\_olid\_offensive & OLID offensive definition & No \\
dspy\_optimized & DSPy-selected few-shot examples & -- \\
dspy\_aligned & DSPy-selected few-shot examples + aligned definition & Yes \\
\bottomrule
\end{tabular}
\end{table}

\subsection{DSPy Automated Optimization}
\label{app:dspy}

We additionally evaluate two automated prompt optimization conditions using DSPy~\citep{khattab2023dspy}, which selects few-shot demonstrations from a labeled validation pool (of 100 examples) to maximize accuracy. \textbf{DSPy Optimized} uses a bare classification signature with no human-written definition (automated counterpart to \texttt{few\_shot}). \textbf{DSPy Aligned} additionally includes the dataset's aligned definition in the instruction field (automated counterpart to \texttt{few\_shot\_aligned\_def}). Optimization is run independently per model--dataset pair.

\subsection{Continuation Prompt}

To measure text-level familiarity (ROUGE-L), we split each text instance at the 40\% mark by word count. We prompt the model with the first 40\% (prefix) using the following template:

\begin{quote}
Continue the following passage faithfully: [PREFIX]
\end{quote}

The model generates a continuation using greedy decoding (temperature 0.0). We then compute the ROUGE-L F1 score between the generated continuation and the remaining 60\% of the ground truth text (suffix).

\paragraph{Additional memorization metrics.}
To confirm that ROUGE-L's limitations as a memorization proxy do not drive our findings, we compute two semantically rich alternatives over the same continuation/suffix pairs. Both use the \texttt{all-MiniLM-L6-v2} encoder---the same encoder as DSF---so any sign difference between memorization and DSF reflects signal type rather than encoder choice. \textbf{Embedding Similarity} is the cosine similarity between the sentence embeddings of the generated continuation and the ground-truth suffix. \textbf{BERTScore}~\citep{zhang2020bertscore} computes token-level F1 via pairwise cosine similarities of contextual token embeddings of the two texts. Both metrics are averaged at the model--dataset level before correlation analysis.

\section{Mixed Effects Regression Models}
\label{app:regression}

We fit four mixed effects logistic regression models to analyze annotation correctness and error correction. All models include message-level random intercepts to account for repeated measures across experimental conditions. Models were estimated using Bayesian variational inference (BinomialBayesMixedGLM in statsmodels).

\paragraph{Model 1: Domain-Alignment Effects}

\textbf{Research Question:} What factors predict correct annotation?

\textbf{Formula:} \texttt{correct $\sim$ C(model) + C(condition) + C(domain) + is\_aligned\_defn + C(domain):is\_aligned\_defn + (1|message)}

\textbf{Reference categories:} Llama-3.1-8B (model), Forum (domain), Misaligned definition (condition)

\begin{table}[H]
\caption{Model 1: Full results for domain-alignment effects.}
\label{tab:model1}
\centering
\small
\begin{tabular}{@{}l d{3} d{2}@{\hspace{4pt}}l@{}}
\toprule
\textbf{Predictor} & \multicolumn{1}{c}{\textbf{Coefficient}} & \multicolumn{2}{c}{\textbf{OR [95\% CI]}} \\
\midrule
Intercept & 1.549 & 4.71 & [4.63, 4.78] \\
\midrule
\multicolumn{4}{@{}l}{\textit{Model Effects (vs. Llama-8B):}} \\
\quad DeepSeek-V3 & 0.499 & 1.65 & [1.58, 1.72] \\
\quad Llama-3.1-70B & 0.237 & 1.27 & [1.22, 1.32] \\
\quad Mistral-7B & 0.384 & 1.47 & [1.42, 1.52] \\
\quad Mistral-Small-24B & 0.326 & 1.39 & [1.33, 1.44] \\
\quad Mixtral-8x7B & 0.456 & 1.58 & [1.52, 1.64] \\
\midrule
\multicolumn{4}{@{}l}{\textit{Domain Effects (vs. Forum):}} \\
\quad Gaming & 2.628 & 13.85 & [13.31, 14.41] \\
\quad News & -0.048 & 0.95 & [0.92, 0.99] \\
\quad Social Media & 1.695 & 5.45 & [5.27, 5.63] \\
\midrule
\multicolumn{4}{@{}l}{\textit{Alignment Effects:}} \\
\quad is\_aligned\_defn & 0.073 & 1.08 & [1.04, 1.11] \\
\quad Gaming $\times$ Aligned & 0.419 & 1.52 & [1.40, 1.65] \\
\quad News $\times$ Aligned & 0.645 & 1.91 & [1.80, 2.02] \\
\quad Social Media $\times$ Aligned & 0.333 & 1.40 & [1.32, 1.47] \\
\midrule
Random Effects SD & 1.156 & 3.18 & [3.11, 3.24] \\
\bottomrule
\end{tabular}
\end{table}

\paragraph{Model 2: Model-Alignment Interaction}

\textbf{Research Question:} Do different models benefit differently from aligned definitions?

\textbf{Formula:} \texttt{correct $\sim$ C(model) + C(condition) + C(domain) + is\_aligned\_defn + C(model):is\_aligned\_defn + (1|message)}

\begin{table}[H]
\caption{Model 2: Model $\times$ Alignment interactions.}
\label{tab:model2}
\centering
\small
\begin{tabular}{@{}lcc@{}}
\toprule
\textbf{Model} & \textbf{Alignment Interaction} & \textbf{OR [95\% CI]} \\
\midrule
Llama-3.1-8B (ref) & 0.195 & 1.22 [1.18, 1.25] \\
Llama-3.1-70B & 0.437 & 1.55 [1.44, 1.67] \\
DeepSeek-V3 & 0.320 & 1.38 [1.28, 1.48] \\
Mistral-Small-24B & 0.202 & 1.22 [1.14, 1.32] \\
Mixtral-8x7B & 0.100 & 1.11 [1.03, 1.19] \\
Mistral-7B & 0.078 & 1.08 [1.01, 1.16] \\
\bottomrule
\end{tabular}
\end{table}

Llama-3.1-70B benefits most from alignment (OR=1.55), despite having the lowest rescue rate. This suggests larger models have stronger priors that are harder to override, but can better utilize guidance when provided.

\paragraph{Model 3: Zero-Shot Predictor}

\textbf{Research Question:} Does zero-shot correctness predict prompted performance?

\textbf{Formula:} \texttt{correct $\sim$ C(model) + C(condition) + C(domain) + zero\_shot\_correct + (1|message)}

\begin{table}[H]
\caption{Model 3: Zero-shot as predictor of prompted correctness.}
\label{tab:model3}
\centering
\small
\begin{tabular}{@{}lcc@{}}
\toprule
\textbf{Predictor} & \textbf{Coefficient} & \textbf{OR [95\% CI]} \\
\midrule
Zero-shot Correct & 1.861 & 6.43 [6.28, 6.58] \\
\midrule
\multicolumn{3}{@{}l}{\textit{Domain Effects (vs. Forum):}} \\
\quad Gaming & 2.124 & 8.36 [8.01, 8.73] \\
\quad News & 0.177 & 1.19 [1.15, 1.24] \\
\quad Social Media & 1.487 & 4.43 [4.26, 4.60] \\
\bottomrule
\end{tabular}
\end{table}

\paragraph{Model 4: Rescue Prediction}

\textbf{Research Question:} Among zero-shot errors, what predicts whether prompting will correct the error?

\textbf{Formula:} \texttt{rescued $\sim$ C(model) + C(condition) + C(domain) + zs\_conf\_std + (1|message)}

This model is fit only on cases where the model was wrong in zero-shot (N=47,672 errors).

\begin{table}[H]
\caption{Model 4: Predicting error correction (rescue) among zero-shot errors.}
\label{tab:model4}
\centering
\small
\begin{tabular}{@{}l d{3} d{2}@{\hspace{4pt}}l@{}}
\toprule
\textbf{Predictor} & \multicolumn{1}{c}{\textbf{Coefficient}} & \multicolumn{2}{c}{\textbf{OR [95\% CI]}} \\
\midrule
ZS Confidence (standardized) & -0.171 & 0.84 & [0.82, 0.87] \\
\midrule
\multicolumn{4}{@{}l}{\textit{Model Effects (vs. Llama-8B):}} \\
\quad DeepSeek-V3 & 0.634 & 1.89 & [1.75, 2.03] \\
\quad Llama-3.1-70B & 0.111 & 1.12 & [1.04, 1.20] \\
\quad Mistral-7B & -0.071 & 0.93 & [0.87, 1.00] \\
\quad Mistral-Small-24B & 0.514 & 1.67 & [1.57, 1.79] \\
\quad Mixtral-8x7B & 0.296 & 1.35 & [1.25, 1.45] \\
\midrule
\multicolumn{4}{@{}l}{\textit{Domain Effects (vs. Forum):}} \\
\quad Gaming & 0.982 & 2.67 & [2.47, 2.89] \\
\quad News & 0.580 & 1.79 & [1.68, 1.89] \\
\quad Social Media & 1.331 & 3.78 & [3.56, 4.02] \\
\bottomrule
\end{tabular}
\end{table}

\section{Additional Experimental Results}
\label{app:performance}

\begin{table}[H]
\caption{Detailed breakdown of model accuracy (\%) under specific misaligned definitions. These values are averaged into the single ``Misaligned'' column in Table~\ref{tab:model_performance}. Columns are ordered by definition scope (Narrow $\to$ Broad). Models below the horizontal rule are not included in regression analyses; Cond.\ AUC is for the six open-weights models.}
\label{tab:full_accuracy}
\centering
\small
\begin{tabular}{@{}lcccccc@{}}
\toprule
& \multicolumn{6}{c}{\textbf{Misaligned Definitions}} \\
\textbf{Model} & \textbf{Twitter} & \textbf{FoxNews} & \textbf{OLID} & \textbf{GameTox} & \textbf{General} & \textbf{Gaming} \\
& \textbf{Hate} & \textbf{Hate} & \textbf{Off.} & \textbf{Tox.} & \textbf{Tox.} & \textbf{Tox.} \\
\midrule
DeepSeek-V3 & 79.2 & 82.0 & 82.6 & 80.8 & 82.0 & 77.4 \\
Mixtral-8x7B & 79.0 & 83.3 & 82.1 & 80.9 & 83.3 & 78.7 \\
Mistral-7B & 79.5 & 83.8 & 84.0 & 80.3 & 84.5 & 76.5 \\
Mistral-Small-24B & 77.2 & 83.1 & 82.9 & 80.3 & 83.5 & 77.3 \\
Llama-3.1-70B & 77.5 & 81.4 & 80.0 & 78.3 & 78.1 & 73.0 \\
Llama-3.1-8B & 77.3 & 79.8 & 78.2 & 76.8 & 76.9 & 72.7 \\
\midrule
GPT-4o-mini & 79.8 & 82.3 & 82.1 & 82.0 & 81.6 & 78.5 \\
Llama-3.3-70B & 78.6 & 83.0 & 81.5 & 77.8 & 78.9 & 74.4 \\
Qwen-2.5-72B & 78.6 & 84.3 & 83.9 & 79.7 & 81.0 & 79.3 \\
\midrule
\textbf{Cond. Avg.} & 78.5 & 82.6 & 81.9 & 79.7 & 81.1 & 76.4 \\
\textbf{Cond. AUC} & .790 & .850 & .867 & .817 & .857 & .841 \\
\bottomrule
\end{tabular}
\end{table}

\begin{table}[H]
\caption{Accuracy (\%) by Dataset and Misaligned Condition. Bold = best for that dataset. ``N/A'' indicates the aligned (non-misaligned) condition.}
\label{tab:dataset_misaligned_acc}
\centering
\small
\begin{tabular}{@{}lcccccc@{}}
\toprule
\textbf{Dataset} & \textbf{foxnews} & \textbf{gametox} & \textbf{gaming} & \textbf{general} & \textbf{olid} & \textbf{twitter} \\
\midrule
FoxNews & N/A & 75.6 & 61.6 & 70.6 & 73.8 & \textbf{78.4} \\
GameTox & 87.2 & N/A & 90.6 & 90.6 & \textbf{90.9} & 85.0 \\
Jigsaw & 75.5 & 75.8 & 70.4 & N/A & \textbf{77.4} & 73.0 \\
OLID & 75.8 & 79.7 & 79.7 & \textbf{79.9} & N/A & 74.3 \\
Twitter & 88.7 & \textbf{89.1} & 80.8 & 85.7 & 84.4 & N/A \\
\bottomrule
\end{tabular}
\end{table}

\begin{table}[H]
\caption{Model sensitivity: best accuracy (across all conditions) minus worst accuracy under misaligned definitions. Models below the horizontal rule are not included in regression analyses.}
\label{tab:model_sensitivity}
\centering
\small
\begin{tabular}{@{}lccc@{}}
\toprule
\textbf{Model} & \textbf{Best Acc.} & \textbf{Worst Acc.} & \textbf{Sensitivity} \\
\midrule
Llama-3.1-70B & 82.1 & 73.0 & 9.1 (highest) \\
Mistral-7B & 84.5 & 76.5 & 8.0 \\
Llama-3.1-8B & 80.1 & 72.7 & 7.4 \\
DeepSeek-V3 & 83.8 & 77.4 & 6.4 \\
Mistral-Small-24B & 83.5 & 77.2 & 6.3 \\
Mixtral-8x7B & 83.3 & 78.7 & 4.6 (lowest) \\
\midrule
Llama-3.3-70B & 83.0 & 74.4 & 8.6 \\
GPT-4o-mini & 84.1 & 78.5 & 5.6 \\
Qwen-2.5-72B & 84.3 & 78.6 & 5.7 \\
\bottomrule
\end{tabular}
\end{table}

\begin{table}[H]
\caption{Corruption rate (\%) by model and misaligned condition. High corruption indicates that correct predictions are frequently flipped to incorrect ones when the definition changes. Models below the horizontal rule are not included in regression analyses.}
\label{tab:corruption_matrix}
\centering
\small
\begin{tabular}{@{}l*{6}{d{1}}@{}}
\toprule
\textbf{Model} & \multicolumn{1}{c}{\textbf{Twitter}} & \multicolumn{1}{c}{\textbf{FoxNews}}
 & \multicolumn{1}{c}{\textbf{OLID}} & \multicolumn{1}{c}{\textbf{GameTox}}
 & \multicolumn{1}{c}{\textbf{General}} & \multicolumn{1}{c}{\textbf{Gaming}} \\
\midrule
DeepSeek-V3 & 14.7 & 10.7 & 6.4 & 7.5 & 7.4 & 12.1 \\
Llama-3.1-70B & 16.8 & 12.3 & 6.0 & 7.5 & 8.3 & 13.3 \\
Llama-3.1-8B & 14.0 & 10.9 & 11.1 & 10.4 & 7.6 & 12.4 \\
Mistral-7B & 8.4 & 4.6 & 4.0 & 8.2 & 2.2 & 9.9 \\
Mistral-Small-24B & 21.0 & 11.3 & 5.0 & 8.2 & 3.1 & 10.0 \\
Mixtral-8x7B & 9.0 & 6.5 & 6.7 & 5.6 & 4.8 & 10.4 \\
\midrule
GPT-4o-mini & 13.2 & 10.3 & 5.7 & 10.9 & 5.7 & 9.0 \\
Llama-3.3-70B & 15.2 & 10.0 & 5.4 & 6.9 & 8.3 & 12.0 \\
Qwen-2.5-72B & 12.6 & 7.2 & 4.6 & 7.4 & 8.6 & 9.6 \\
\bottomrule
\end{tabular}
\end{table}

\begin{table}[H]
\caption{Rescue rate by model on the Jigsaw Unintended Bias dataset~\citep{borkan2019nuanced}, which was designed to reduce identity-group annotation bias. Rescue rate is $P(\text{correct} \mid \text{prompted}, \text{zero-shot wrong})$, computed over all non-zero-shot conditions (aligned definition, few-shot, few-shot+definition, DSPy Optimized, DSPy Aligned, and five misaligned conditions), matching the methodology of Table~\ref{tab:rescue_by_model}. All models remain well below 50\%, confirming that decision stickiness is not an artifact of Jigsaw's known label-bias limitations.}
\label{tab:jigsaw_rescue}
\vskip 0.1in
\centering
\small
\begin{tabular}{@{}lccc@{}}
\toprule
\textbf{Model} & \textbf{Rescue Rate} & \textbf{95\% CI} & \textbf{N (ZS err.\ obs.)} \\
\midrule
DeepSeek-V3       & 38.2\% & [36.4, 40.1] & 2,640 \\
Mistral-Small-24B & 35.6\% & [33.9, 37.4] & 2,810 \\
Llama-3.1-8B      & 34.3\% & [32.5, 36.0] & 2,790 \\
GPT-4o-mini       & 31.4\% & [29.6, 33.3] & 2,350 \\
Mixtral-8x7B      & 31.0\% & [29.3, 32.8] & 2,790 \\
Llama-3.1-70B     & 30.8\% & [29.1, 32.5] & 2,870 \\
Llama-3.3-70B     & 28.2\% & [26.6, 29.9] & 2,790 \\
Mistral-7B        & 27.8\% & [26.1, 29.5] & 2,680 \\
Qwen-2.5-72B      & 24.9\% & [23.2, 26.7] & 2,330 \\
\bottomrule
\end{tabular}
\vskip -0.1in
\end{table}

%% Iterative, multi-turn rescure

\subsection{Multi-Turn Rescue Experiment}
\label{app:multiturn}
A natural follow-up to the one-shot rescue analysis (Sec.~\ref{sec:rq2}) is whether
iterative, history-aware prompting can break decision stickiness. For each
model--dataset pair we sampled 100 zero-shot errors and applied a 3-step
rescue sequence: Turn 1 adds a few labeled examples, Turn 2 adds the aligned
task definition, and Turn 3 asks the model to reconsider. At each turn the
model is shown its previous answers and confidence scores, so later prompts
see the full history.

Table~\ref{tab:multiturn_rescue} reports cumulative rescue rates by model, both
overall and restricted to high-confidence errors. Iterative prompting helps
modestly --- particularly when the aligned definition is added at Turn 2 ---
but rescue plateaus well below the one-shot rate of 34.8\% reported in
Table~\ref{tab:rescue_by_model}, and high-confidence errors remain almost
entirely uncorrected (8.5\% rescued after three turns).

% \textit{Caveat.} Llama-3.1-8B-Instruct refused to provide a classification in
% later turns on 78.8\% of sampled items; we drop those instances before
% computing its rates and interpret its row cautiously.

\begin{table}[H]
\caption{Cumulative rescue rates over three multi-turn correction attempts.
``High-conf'' columns restrict to zero-shot errors with confidence $>0.9$.}
\label{tab:multiturn_rescue}
\centering
\small
\begin{tabular}{@{}l*{6}{d{1}}@{}}
\toprule
\textbf{Model} & \multicolumn{1}{c}{\textbf{Rescue@1}} & \multicolumn{1}{c}{\textbf{Rescue@2}}
 & \multicolumn{1}{c}{\textbf{Rescue@3}} & \multicolumn{1}{c}{\textbf{High-conf.@1}}
 & \multicolumn{1}{c}{\textbf{High-conf.@2}} & \multicolumn{1}{c}{\textbf{High-conf.@3}} \\
\midrule
DeepSeek-V3 & 8.2 & 20.8 & 27.4 & 0.0 & 5.1 & 6.8 \\
Llama-3.1-70B & 10.6 & 14.6 & 19.0 & 2.9 & 5.0 & 8.0 \\
Llama-3.1-8B$^{\dagger}$ & 21.7 & 35.8 & 35.8 & 0.0 & 16.7 & 16.7 \\
Mistral-7B & 3.0 & 12.3 & 12.7 & 1.2 & 8.5 & 9.1 \\
Mistral-Small-24B & 7.0 & 12.4 & 12.9 & 4.1 & 9.6 & 9.6 \\
Mixtral-8x7B & 5.8 & 17.1 & 18.5 & 1.9 & 6.5 & 7.4 \\
\midrule
\textbf{Pooled} & 7.5 & 16.3 & 18.7 & 2.0 & 7.1 & 8.5 \\
\bottomrule
\end{tabular}\\
\vspace{0.3em}
\footnotesize $^{\dagger}$78.8\% of sampled items dropped due to classification refusals.
\end{table}

\section{Familiarity Analysis Details}
\label{app:familiarity}

\subsection{DSF Robustness Across Embedding Models}
\label{app:dsf_embeddings}

The DSF score is a cosine similarity in an embedding space, so its value depends on the encoder used. To rule out that our results are an artifact of choosing any single embedding model, we recompute DSF using six diverse sentence-embedding models spanning small (MiniLM, mpnet) to large (bge-large, e5-large, instructor-large) open-source encoders, plus a proprietary OpenAI model (text-embedding-3-small). The full model identifiers are:
\begin{itemize}[nosep,leftmargin=*]
  \item MiniLM: \texttt{sentence-transformers/all-MiniLM-L6-v2}
  \item MPNet: \texttt{sentence-transformers/all-mpnet-base-v2}
  \item BGE-large: \texttt{BAAI/bge-large-en-v1.5}
  \item E5-large: \texttt{intfloat/e5-large-v2}
  \item Instructor-large: \texttt{hkunlp/instructor-large}
  \item OpenAI: \texttt{text-embedding-3-small}
\end{itemize}

We then define \emph{consensus DSF} for each model--dataset cell as the unweighted mean DSF across the six encoders, and adopt it as the primary DSF metric in the main text.

The DSF analysis covers all nine evaluated models across six datasets (the five primary datasets plus Jigsaw Unintended Bias~\citep{borkan2019nuanced}), yielding $N=54$ model--dataset pairs. Model performance is reported in Table~\ref{tab:model_performance}.

\paragraph{Per-embedding partial correlations.}
Table~\ref{tab:dsf_by_embed} reports the partial correlation between DSF and zero-shot accuracy (controlling for dataset) separately for each embedding choice. All six embeddings yield a positive partial correlation, ranging from $+0.30$ to $+0.49$. Larger encoders (bge-large, e5-large, OpenAI) tend to yield slightly stronger correlations than the compact MiniLM. The consensus row ($+0.41$) sits at the upper end of this range, indicating that averaging across encoders denoises rather than dilutes the signal.

\begin{table}[H]
\caption{DSF--accuracy correlations across six sentence-embedding models on the extended $N=54$ panel (zero-shot accuracy, partial correlation controls for dataset). Consensus DSF is the per-cell mean across embeddings, and is the metric reported in the main text.}
\label{tab:dsf_by_embed}
\centering
\small
\begin{tabular}{@{}lcc@{}}
\toprule
\textbf{Embedding model} & \textbf{Raw $r$} & \textbf{Partial $r$} \\
\midrule
all-MiniLM-L6-v2          & $+0.73$ & $+0.36$ \\
all-mpnet-base-v2         & $+0.77$ & $+0.33$ \\
BAAI/bge-large-en-v1.5    & $+0.74$ & $+0.49$ \\
intfloat/e5-large-v2      & $+0.60$ & $+0.43$ \\
hkunlp/instructor-large   & $+0.72$ & $+0.30$ \\
OpenAI text-embedding-3-small & $+0.72$ & $+0.43$ \\
\midrule
\textbf{Consensus (mean across 6)} & $\mathbf{+0.74}$ & $\mathbf{+0.41}$ \\
\bottomrule
\end{tabular}
\end{table}

\paragraph{Pairwise embedding agreement.}
DSF values are highly concordant across embedding choices. Table~\ref{tab:dsf_pairwise} reports the Pearson correlation between DSF score vectors produced by different embedding models on the $N=54$ panel; all pairwise correlations exceed $0.87$, and most exceed $0.95$. This indicates that the variation in Table~\ref{tab:dsf_by_embed} reflects what each embedding emphasizes (e.g., lexical vs. semantic, general vs. domain-tuned) rather than disagreement about which model--dataset pairs are high- vs. low-DSF.

\begin{table}[H]
\caption{Pairwise Pearson correlations between DSF score vectors across six embedding models, on the $N=54$ panel. All embeddings agree closely on the relative ordering of model--dataset cells.}
\label{tab:dsf_pairwise}
\centering
\scriptsize
\begin{tabular}{@{}lcccccc@{}}
\toprule
 & MiniLM & mpnet & bge & e5 & instr. & OpenAI \\
\midrule
MiniLM      & $1.00$ & $0.97$ & $0.98$ & $0.93$ & $0.97$ & $0.97$ \\
mpnet       &        & $1.00$ & $0.96$ & $0.88$ & $0.97$ & $0.97$ \\
bge-large   &        &        & $1.00$ & $0.95$ & $0.98$ & $0.99$ \\
e5-large    &        &        &        & $1.00$ & $0.94$ & $0.93$ \\
instructor  &        &        &        &        & $1.00$ & $0.98$ \\
OpenAI      &        &        &        &        &        & $1.00$ \\
\bottomrule
\end{tabular}
\end{table}

\paragraph{Takeaway.}
The DSF--accuracy association is not an artifact of a particular encoder. Across six embedding models the partial correlation remains positive and meaningful, and the embeddings closely agree on which model--dataset pairs score high or low. We therefore report consensus DSF in the main text as a stable, encoder-agnostic operationalization of the metric.

\subsection{Min-K\% Prob as a Memorization Baseline}
\label{app:mink}

Min-K\% Prob~\citep{shi2024detecting} computes the average log probability of the $k\%$ least likely tokens in a text. Lower (more negative) values indicate the model finds certain tokens surprising, suggesting the text was \emph{not} memorized during training. Computing Min-K\% requires per-token log probabilities, which are unavailable for most frontier API models in our evaluation. We therefore use ROUGE-L as the primary memorization baseline, as it is computable via text generation for any model.

To verify that this choice does not change our qualitative conclusions, we applied Min-K\% ($k=20\%$) to the two open-weights models in our panel for which log probabilities are available (Llama-3.1-8B and Mistral-7B, base versions) across the five paper datasets ($N=10$ model--dataset pairs). Table~\ref{tab:mink_comparison} reports mixed-effects model coefficients (Accuracy $\sim$ metric + $(1\mid\text{dataset})$ + $(1\mid\text{model})$) for ROUGE-L, Min-K\%, and DSF on this subset. Min-K\% and ROUGE-L are directionally consistent---both show negative standardized coefficients while DSF remains positive---confirming that our conclusions do not depend on the choice of memorization metric.

\begin{table}[H]
\caption{Mixed-effects model coefficients for memorization metrics and DSF on the $N=10$ subset (Llama-3.1-8B and Mistral-7B $\times$ 5 paper datasets) where Min-K\% Prob is computable. Standardized coefficients allow cross-metric comparison of effect size.}
\label{tab:mink_comparison}
\centering
\small
\begin{tabular}{@{}lrrr@{}}
\toprule
\textbf{Metric} & \textbf{Std.\ Coef.} & \textbf{Raw Coef.} & \textbf{$p$} \\
\midrule
Text Memorization (ROUGE-L)   & $-0.755$ & $-2.479$ & $0.001$ \\
Memorization (Min-K\% Prob)   & $-0.596$ & $-0.044$ & $0.036$ \\
Definition Familiarity (DSF)  & $+0.838$ & $+0.425$ & $<0.001$ \\
\bottomrule
\end{tabular}
\end{table}

\subsection{Semantic Memorization Metrics: BERTScore and Embedding Similarity}
\label{app:semantic_memorization}

ROUGE-L captures lexical overlap via the longest common subsequence and may therefore underestimate memorization when a model reproduces the \emph{content} of a passage with paraphrased surface form.  We additionally evaluate two semantically richer alternatives that operate on the same generated continuations as ROUGE-L, so the comparison isolates the scoring function rather than the generation procedure.

\paragraph{BERTScore.}
BERTScore~\citep{zhang2020bertscore} replaces $n$-gram matching with contextual token-embedding similarity. Each token in the generated continuation $\hat{y}$ and the reference suffix $y$ is encoded with a pretrained contextual encoder, and tokens are greedily aligned by maximum cosine similarity:
\begin{equation*}
R_{\text{BERT}} = \frac{1}{|y|}\sum_{y_i \in y}\max_{\hat{y}_j \in \hat{y}} \mathbf{y}_i^\top \hat{\mathbf{y}}_j,
\quad
P_{\text{BERT}} = \frac{1}{|\hat{y}|}\sum_{\hat{y}_j \in \hat{y}}\max_{y_i \in y} \mathbf{y}_i^\top \hat{\mathbf{y}}_j,
\end{equation*}
with $F_{1,\text{BERT}}$ as their harmonic mean. Because the embeddings are contextual, BERTScore credits paraphrases and near-synonyms that ROUGE-L would penalize. We use \texttt{microsoft/deberta-xlarge-mnli} as the encoder (the default recommended by~\citet{zhang2020bertscore} for English) and report $F_{1,\text{BERT}}$ as the primary score.

\paragraph{Embedding cosine similarity.}
Embedding similarity moves one level higher, comparing whole-sentence representations rather than token alignments:
\begin{equation*}
\text{EmbSim}(\hat{y}, y) \;=\; \frac{\phi(\hat{y})^\top \phi(y)}{\|\phi(\hat{y})\|\,\|\phi(y)\|},
\end{equation*}
where $\phi(\cdot)$ is a sentence encoder. This is the loosest of the three notions of textual overlap: a continuation that captures the gist of the reference but rewords it freely can still score high. To ensure that any difference between EmbSim and DSF reflects \emph{what} is being measured (continuation--reference alignment vs. definition--self-description alignment) rather than \emph{which encoder} does the measuring, we use the same \texttt{all-MiniLM-L6-v2} encoder as in DSF.

\paragraph{Spectrum of memorization sensitivity.}
Taken together, ROUGE-L, BERTScore, and EmbSim form a progression from purely lexical to fully semantic notions of reproduction. If text memorization were driving annotation accuracy and ROUGE-L were simply too crude to detect it, we would expect the correlation with accuracy to strengthen (or at least become positive) as we relax the lexical constraint. Table~\ref{tab:familiarity_correlations} shows the opposite: all three text-memorization metrics yield negative partial correlations with accuracy after controlling for dataset, while DSF remains positive. The negative direction is preserved across the full lexical--semantic spectrum, indicating that the issue is not ROUGE-L's lexical sensitivity but that text reproduction itself, however measured, does not predict annotation performance.

\subsection{Definition-Specific Familiarity (DSF) Variants}

\begin{figure}[H]
\centering
\includegraphics[width=\textwidth]{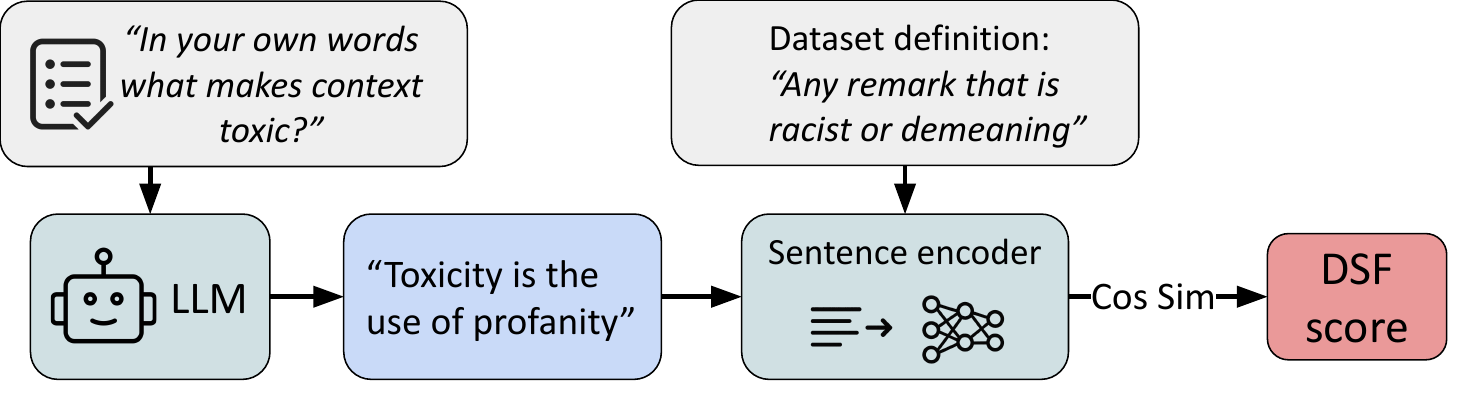}
\caption{\textbf{Conceptual overview of Definition-Specific Familiarity (DSF).} We first ask the model to explain the target concept in its own words. We treat the model-generated explanation as an operational proxy for its internal concept. We then encode the explanation and the dataset's reference definition with the same sentence encoder and compute their cosine similarity. A higher DSF score indicates closer semantic alignment between the model's elicited concept and the definition expected by the dataset. A lower score indicates poorer alignment.}
\label{fig:conceptual_dsf}
\end{figure}

In the main text, we define \textbf{Definition-Specific Familiarity (DSF)} as the cosine similarity between the embedding of the dataset's ground-truth definition and the model's own description of the concept. Figure~\ref{fig:conceptual_dsf} illustrates this primary metric; below, we detail it alongside two alternative variants we explored.

\paragraph{DSF (Primary Metric).}
This is the metric reported in the main results. We prompt the model with a neutral instruction to describe the concept: \textit{``You are documenting your internal understanding of a labeling concept. Concept: [Concept Name]. Describe, in your own words, what content qualifies as this concept.''} We then compute the cosine similarity between the embedding of this generated response and the embedding of the dataset's full definition. We found this approach correlated most strongly with model performance.

\paragraph{Alternative Variant: Positive DSF.}
This variant isolates the ``positive'' class definition (e.g., what \textit{is} toxicity) from the ``negative'' class. It uses \textbf{multi-prompt elicitation} to reduce variance, averaging similarities across three prompt styles: \textit{Direct} (``You are documenting...''), \textit{Expert} (``As an expert annotator...''), and \textit{Examples} (``Without examples, describe...'').

\paragraph{Alternative Variant: Enhanced DSF.}
This metric extends Positive DSF by adding \textbf{domain contextualization} (e.g., \textit{``in online gaming chat''} for GameTox) and \textbf{negative class alignment} (comparing the model's description of what the concept is \textit{not} to the dataset's negative definition). The final score is the average of the positive and negative alignment scores.

\begin{table}[H]
\caption{Familiarity metrics by dataset (averaged across the six open-weights models, zero-shot). DSF is reported as consensus DSF across six sentence encoders. Jigsaw has the lowest DSF and is the hardest dataset; GameTox has the highest accuracy with negligible text memorization, illustrating that conceptual alignment, not memorization, tracks performance.}
\label{tab:familiarity_dataset}
\centering
\small
\begin{tabular}{@{}lcccc@{}}
\toprule
\textbf{Dataset} & \textbf{Domain} & \textbf{ROUGE-L} & \textbf{Consensus DSF} & \textbf{Accuracy} \\
\midrule
Twitter & Social Media & 0.045 & 0.804 & 0.901 \\
GameTox & Gaming       & 0.001 & 0.784 & 0.902 \\
OLID    & Social Media & 0.063 & 0.710 & 0.751 \\
FoxNews & News         & 0.061 & 0.652 & 0.723 \\
Jigsaw  & Forum        & 0.084 & 0.458 & 0.705 \\
\bottomrule
\end{tabular}
\end{table}

\section{Generalization Beyond Toxicity}
\label{app:generalization}

To probe whether our two main findings---the positive DSF--accuracy association (RQ1) and decision stickiness (RQ2)---extend beyond safety tasks, we conduct supplementary experiments on two binary classification tasks: \emph{irony detection} (SemEval-2018 Task 3; \citealt{vanhee2018semeval}) and \emph{subjectivity detection} \citep{pang2004sentimental}. Both use explicit human-authored definitions and binary labels, making them structurally comparable to the toxicity setup. We evaluate the same 9 models from Table~\ref{tab:models} on 1{,}000 instances per dataset--model pair. Because this extension contains only two datasets and fewer prompting conditions, we treat it as a generalization check rather than evidence of universality.

\paragraph{DSF extends directionally beyond toxicity.}
Across the 18 model--dataset pairs, DSF is positively associated with zero-shot accuracy after controlling for dataset (partial $r=+0.343$). This matches the direction of the main result, suggesting that DSF captures model--task fit beyond toxicity, although the two-dataset sample makes this evidence necessarily preliminary.

\paragraph{Decision stickiness persists beyond toxicity.}
Few-shot prompting rescues $45.0\%$ ($95\%$ CI $[43.2\%, 46.8\%]$) of $3{,}052$ zero-shot errors, compared with $34.8\%$ on the toxicity tasks (Table~\ref{tab:rescue_by_model}). Thus, even where examples help more, most initial errors remain uncorrected. Tables~\ref{tab:nonsafety_accuracy} and~\ref{tab:nonsafety_rescue} report condition-level accuracy and per-model rescue rates.

\begin{table}[H]
\caption{Accuracy (\%) by condition on the two non-safety datasets (irony and subjectivity), averaged across the two datasets. Sampling and inference settings match the main experiments.}
\label{tab:nonsafety_accuracy}
\centering
\small
\begin{tabular}{@{}lccc@{}}
\toprule
\textbf{Model} & \textbf{Zero-Shot} & \textbf{Aligned Def} & \textbf{Few-Shot} \\
\midrule
Llama-3.3-70B     & 73.3 & 70.4 & 86.2 \\
Mixtral-8x7B      & 71.2 & 76.1 & 84.2 \\
Llama-3.1-70B     & 71.0 & 68.9 & 85.2 \\
Mistral-7B        & 68.8 & 65.9 & 73.9 \\
DeepSeek-V3       & 68.1 & 72.4 & 76.9 \\
Mistral-Small-24B & 63.2 & 71.8 & 79.3 \\
GPT-4o-mini       & 62.7 & 73.7 & 81.9 \\
Qwen-2.5-72B      & 57.1 & 60.7 & 69.7 \\
Llama-3.1-8B      & 50.2 & 57.1 & 72.1 \\
\midrule
\textbf{Cond. Avg} & 65.1 & 68.5 & 78.8 \\
\bottomrule
\end{tabular}
\end{table}

\begin{table}[H]
\caption{Rescue rate by model in the few-shot condition on the two non-safety datasets (irony and subjectivity). Rescue rate is $P(\text{correct} \mid \text{few-shot}, \text{zero-shot wrong})$, computed pooled across the two datasets.}
\label{tab:nonsafety_rescue}
\centering
\small
\begin{tabular}{@{}lccc@{}}
\toprule
\textbf{Model} & \textbf{Rescue Rate} & \textbf{95\% CI} & \textbf{N (ZS err.)} \\
\midrule
Mixtral-8x7B      & 59.9\% & [53.9, 65.7] & 277 \\
Llama-3.1-8B      & 52.8\% & [48.3, 57.3] & 498 \\
Llama-3.1-70B     & 52.8\% & [46.8, 58.6] & 290 \\
GPT-4o-mini       & 52.3\% & [47.1, 57.4] & 373 \\
Llama-3.3-70B     & 50.6\% & [44.4, 56.7] & 267 \\
Mistral-Small-24B & 47.1\% & [41.9, 52.4] & 367 \\
DeepSeek-V3       & 37.6\% & [32.3, 43.2] & 319 \\
Qwen-2.5-72B      & 27.9\% & [23.6, 32.6] & 401 \\
Mistral-7B        & 21.5\% & [16.7, 27.0] & 260 \\
\midrule
\textbf{Overall}  & 45.0\% & [43.2, 46.8] & 3{,}052 \\
\bottomrule
\end{tabular}
\end{table}

\paragraph{Stance detection: a structurally different task.}
As a further check, we ran a preliminary experiment on stance detection (SemEval-2016 Task 6; \citealt{mohammad2016semeval}) across 5 targets and 8 models. The DSF direction remains positive (partial $r=+0.295$, $N=40$ target--model pairs). Stance detection differs structurally from our other tasks, however: it requires choosing between two explicit alternatives (``in favor'' / ``against'') rather than classifying content against a single concept definition, which may introduce additional sources of variation beyond conceptual alignment around the definition itself. We therefore treat stance as directionally consistent but not as clean a test of the DSF--accuracy relationship as irony and subjectivity, and we report it here only as supporting evidence.

\paragraph{Takeaway.}
Irony and subjectivity show the same qualitative pattern as the primary experiments: better definition alignment accompanies higher accuracy, while most initial errors resist correction. These results broaden the evidence beyond toxicity, but the larger and more controlled toxicity experiments remain the basis of our main claims.

\section{Model and Dataset Use Across Experiments}
\label{app:coverage}

Our experiments use a two-tier structure for both models and datasets, summarized in Table~\ref{tab:coverage}.

The main-paper descriptive RQ2/RQ3 result tables and figures (Tables~\ref{tab:rescue_by_model},~\ref{tab:misalignment_results} and Figs.~\ref{fig:rescue_confidence},~\ref{fig:calibration_curve}) report values aggregated across all nine evaluated models (Llama-3.1-8B, Llama-3.1-70B, Llama-3.3-70B, Mistral-7B, Mistral-Small-24B, Mixtral-8x7B, DeepSeek-V3, GPT-4o-mini, Qwen-2.5-72B). All mixed-effects regression analyses (Table~\ref{tab:glm_results} in the main paper and Tables~\ref{tab:model1}--\ref{tab:model4} in the appendix) and the multi-turn rescue analysis (Table~\ref{tab:multiturn_rescue}) are scoped to the original six open-weights models (Llama-3.1-8B, Llama-3.1-70B, Mistral-7B, Mistral-Small-24B, Mixtral-8x7B, DeepSeek-V3) to keep statistical inference tied to the original six-model design; GPT-4o-mini, Llama-3.3-70B, and Qwen-2.5-72B contribute to the descriptive and performance tables but are excluded from the mixed-effects regressions.

The dataset side follows the same pattern. The five primary toxicity datasets (Twitter Hate, OLID, GameTox, Fox News, Jigsaw Toxic Comments) are the analytic core. The three supplementary datasets---Jigsaw Unintended Bias, SemEval Irony, and Subjectivity---serve targeted checks: Jigsaw Unintended Bias tests robustness to label-bias concerns, while irony and subjectivity probe whether the RQ1 and RQ2 patterns extend beyond toxicity (Appendix~\ref{app:generalization}).

\begin{table}[H]
\caption{Coverage by analysis. ``6 core'' and ``9 models'' denote the model tiers defined above; ``5 main'' denotes the primary toxicity datasets; Jigsaw UB = Jigsaw Unintended Bias.}
\label{tab:coverage}
\centering
\small
\begin{tabular}{@{}p{6.2cm}p{8.5cm}@{}}
\toprule
\textbf{Coverage} & \textbf{Where it applies} \\
\midrule
9 models $\times$ 5 main datasets & Tables~\ref{tab:rescue_by_model},~\ref{tab:misalignment_results}; Figs.~\ref{fig:rescue_confidence},~\ref{fig:calibration_curve}; Table~\ref{tab:model_performance}; Tables~\ref{tab:full_accuracy},~\ref{tab:model_sensitivity},~\ref{tab:corruption_matrix} (extended models shown below midrule) \\
6 core models $\times$ 5 main datasets & Tables~\ref{tab:glm_results},~\ref{tab:model1}--\ref{tab:model4},~\ref{tab:dataset_misaligned_acc},~\ref{tab:multiturn_rescue},~\ref{tab:familiarity_dataset} \\
9 models $\times$ 5 main + Jigsaw UB ($N{=}54$) & Tables~\ref{tab:familiarity_correlations},~\ref{tab:dsf_by_embed},~\ref{tab:dsf_pairwise} (familiarity / DSF) \\
9 models $\times$ Jigsaw UB only & Table~\ref{tab:jigsaw_rescue} (rescue-rate robustness check) \\
9 models $\times$ Irony + Subjectivity & Tables~\ref{tab:nonsafety_accuracy},~\ref{tab:nonsafety_rescue} (non-safety generalization) \\
Llama-3.1-8B + Mistral-7B (base) $\times$ 5 main & Table~\ref{tab:mink_comparison} (Min-K\%, requires logprob access) \\
\bottomrule
\end{tabular}
\end{table}

\end{document}